\documentclass[preprint,11pt]{elsarticle}

\usepackage{amssymb}
\usepackage{booktabs}
\usepackage{amsthm}
\usepackage{mathtools}
\usepackage{algorithm}
\usepackage{algpseudocode}
\usepackage{caption}
\usepackage{subcaption}
\usepackage[top=2.5cm,bottom=2.0cm,left=2.0cm,right=2.5cm]{geometry}
\usepackage{enumitem} 
\usepackage[flushleft]{threeparttable}
\usepackage{upgreek}
\usepackage{lineno,hyperref}
\modulolinenumbers[1]

\journal{Reliability Engineering \& System Safety}

\begin{document}

\begin{frontmatter}


\title{Inference and dynamic decision-making for deteriorating systems \\ with probabilistic dependencies through Bayesian networks \\ and deep reinforcement learning}



\author[1]{P.G. Morato\corref{cor1}}
\ead{pgmorato@uliege.be}
\author[2]{C.P. Andriotis}
\author[3]{K.G. Papakonstantinou}
\author[1]{P. Rigo}

\cortext[cor1]{Corresponding author}
\address[1]{ANAST, Department of ArGEnCo, University of Liege, 4000, Liege, Belgium}
\address[2]{Faculty of Architecture and the Built Environment, Delft University of Technology, 2628 BL Delft, The Netherlands}
\address[3]{Department of Civil \& Environmental Engineering, The Pennsylvania State University, University Park, PA 16802, USA \vspace{-0.4cm}}

\begin{abstract}
In the context of modern engineering, environmental, and societal concerns, there is an increasing demand for methods able to identify rational management strategies for civil engineering systems, minimizing structural failure risks while optimally planning inspection and maintenance (I\&M) processes. Most available methods simplify the I\&M decision problem to the component level, often assuming statistical, structural, or cost independence among components, due to the computational complexity associated with global optimization methodologies under joint system-level state descriptions. In this paper, we propose an efficient algorithmic framework for inference and decision-making under uncertainty for engineering systems exposed to deteriorating environments, providing optimal management strategies directly at the system level. In our approach, the decision problem is formulated as a factored partially observable Markov decision process, whose dynamics are encoded in Bayesian network conditional structures. The methodology can handle environments under equal or general, unequal deterioration correlations among components, through Gaussian hierarchical structures and dynamic Bayesian networks, decoupling the originally joint system state space to component networks conditional on shared random variables. In terms of policy optimization, we adopt a deep decentralized multi-agent actor-critic (DDMAC) reinforcement learning approach, in which the policies are approximated by actor neural networks guided by a critic network. By including deterioration dependence in the simulated environment, and by formulating the cost model at the system level, DDMAC policies intrinsically consider the underlying system-effects. This is demonstrated through numerical experiments conducted for both a 9-out-of-10 system and a steel frame under fatigue deterioration. Results demonstrate that DDMAC policies offer substantial benefits when compared to state-of-the-art heuristic approaches. The inherent consideration of system-effects by DDMAC strategies is also interpreted based on the learned policies.
\end{abstract}




\end{frontmatter}


\section{Introduction}\label{Sec:intro}
Managing engineering systems, by controlling the risks of adverse events and optimally allocating inspection and repair resources, is crucial for securing societal progress, improving the quality of life at the community level and maximizing economic returns from an individual perspective \cite{rackwitz2005socio, Faber2003RiskIntroduction}. Research efforts devoted to the development of risk-based inspection and maintenance planning methods have increased considerably during the last decade \cite{Frangopol2016Life-cycleDirections, frangopol2004probabilistic, frangopol2007maintenance}. Increasing societal consciousness on sustainability, along with the expanding wealth of data from our structural systems and infrastructure, require, and enable, more efficient management policies \cite{Frangopol2016Life-cycleDirections}. Such policies need to support decision-making, for both newly designed systems and existing ones, through life-cycle plans that integrally account for interventions (e.g., repairs, retrofits, etc.) and data collection methods (e.g., inspections, structural health monitoring, etc).     

Most available inspection and maintenance (I\&M) planning methods assume independence among the constitutive components, primarily driven by the practical need to tame the involved computational complexities associated with solving such a decision-making optimization problem under uncertainty \cite{Straub2004Thesis, StraubDBN2009}. At the component level, existing risk-based I\&M methods can be classified according to their capabilities of modeling physically-based deterioration processes, e.g., fatigue or corrosion deterioration \cite{papakonstantinou2013probabilistic, Lotsberg2016ProbabilisticStructures, yang2018probabilistic}, and depending on their policy optimization approaches, namely, static decision rules, adaptive decision rules prescribed by heuristics or adaptive decision rules defined as a function of the dynamically updated history of actions and observations. 

Some methods focus on the optimization of predefined static decision rules, planning inspections at equidistant intervals or when a prescribed failure probability threshold is surpassed, and prescribing maintenance interventions if a certain damage indicator is observed, e.g., crack detection \cite{Straub2004Thesis, nielsen2018computational, Long2018TheInformation, hlaing2020effect}. While these approaches can provide reasonable and effective policies in some specific scenarios, the optimality of the policies depends greatly on the designer's experience when defining the heuristic combinations for the policy search, since they cannot consider all policies within the vast available policy space, which could in turn result more optimal than the originally considered predefined heuristics \cite{morato2022optimal, hlaing2022inspection}. In other existing methods, while inspection planning decision rules are defined a priori, the maintenance policy is adaptive, properly updating the involved thresholds based on new information \cite{bismut2018adaptive,bismut2021optimal}. In these cases, action planning is formulated based on optimization techniques which need to be repeated for the number of all desired updates. While still operating on a limited policy subspace, such sophisticated methods correctly identify the need to go beyond static thresholds and accordingly provide solution approaches.

Methods based on Markov Decision Processes (MDPs) and Partially Observable MDPs (POMDPs) aim, on the other hand, at addressing the problem in a global optimization sense, outside the limitations of threshold-based or time-based formulations. Early works on the application of Markov decision processes for managing deteriorating engineering cases include \cite{Corotis_POMDP_2005, Papakonstantinou2014Part1, Papakonstantinou2014Part2, papakonstantinou2014optimum}. Founded on the principled mathematical properties offered by dynamic programming, either under full or partial observability, additional formulations have been proposed, e.g., in \cite{andriotis2021value, Pozzi_Hierarchical_2016, POzziMilad2015}. In the same class of applications, a recent POMDP-based approach proposed in \cite{morato2022optimal} demonstrated that POMDP policies outperform heuristic-based policies, as exemplified in physically-based numerical examples featuring fatigue deterioration. POMDP policies are defined as a function of the belief about the condition states, i.e., the probability distribution over states, which is a sufficient statistic of the prior history of actions and observations, recursively encoding it through forward Bayesian updates.

As mentioned before, many of the existing I\&M methods formulate the decision-making problem at the component level. However, disregarding the essential interrelations among the system constituents, although allowing for a substantial simplification of the decision-making problem, may result in sub-optimal and even non-conservative policies in some cases. The need for I\&M methods capable of determining policies at the system level has long been identified by the risk research community. Early works approaching the problem at the system level include \cite{Thoft-Christensen1982, ito1992non, deodatis1996reliability, enevoldsen1993reliability}. In \cite{Straub2015}, the fatigue details were classified according to the fatigue design factor, establishing a simplified approach for identifying system policies. More recently, \cite{LuqueDBN2019,DBNLuque2016} proposed a static I\&M planning optimization relying on dynamic Bayesian networks to efficiently model deterioration, cost and reliability dependence among the structural elements. In this method, the policy is computed by optimizing static heuristic decision rules, with decision variables including, among others, equidistant inspections, number of inspected components, component prioritization, and repair thresholds based on observations. As with all static policy optimization methods, explained before, the policies are constrained to the set of predefined heuristic rules, out of the immense space of possible policies, which is substantially enlarged now in structural system settings. 

Addressing the important complexities of managing large engineering systems, a deep reinforcement learning (DRL) method has been introduced in \cite{Andriotis2019ManagingLearning}, motivated by the success of deep reinforcement learning algorithms in complex game environments, e.g., in \cite{silver2016mastering, silver2017mastering, mnih2013playing}. In particular, a multi-agent actor-critic DRL scheme is developed in \cite{Andriotis2019ManagingLearning}, relying on (PO)MDPs for simulating the deteriorating environment, and demonstrating the capabilities of deep reinforcement learning approaches for identifying optimal policies in vast high-dimensional state, action and observation spaces. Thereafter, a modified version of this method has also been applied for solving system I\&M decision-making problem under constraints, e.g., imposed risk thresholds or budget limitations \cite{andriotis2021deep}. In general, DRL approaches offer computational benefits in high-dimensional state spaces, mitigating the need for exhaustive state exploration by leveraging a function parametrization over the state space \cite{wei2020optimal,fan2022systematic,yang2022deep}.

In this work, we formalize an efficient modeling framework for inference and decision-making under uncertainty for engineering systems, directly generating management strategies at the system level. In terms of inference, and addressing the general computational challenges associated with probabilistic analysis of multi-component engineering systems, our proposed methodology builds on top of adept Dynamic Bayesian Network (DBN) formulations \cite{StraubDBN2009,DBNLuque2016}, modeling environments described by deterioration dependencies among components through Gaussian hierarchical structures, with the objective of decoupling the joint system space to independent component networks conditional on common influencing random variables. This decomposition results in a linear computational complexity with the number of components that otherwise increases exponentially in the joint system space. Furthermore, in this paper, the Gaussian hierarchical model is originally expanded and enhanced to enable the treatment of general, unequal deterioration correlation scenarios and dependence alterations after a maintenance action is taken for some of the components. In our developed framework, the transitional probabilistic model should appropriately consider the common random variables, and thus the algorithmic steps for properly updating the belief state under deterioration correlation within the DBN framework are described in detail. 

While our developed generalized inference framework is applicable regardless of the decision-making method used for the generation of management policies, by formulating the decision-making problem as a factored POMDP, whose dynamics are encoded as Bayesian network conditional structures, the aforementioned efficient, modeling and inference framework can be seamlessly and naturally integrated with sophisticated decision-making optimization methods. In this regard, we adopt a deep decentralized multi-agent actor-critic (DDMAC) scheme, in which the system policies are approximated by actor neural networks, at a component level, guided via system level value function estimates approximated by a critic network \cite{Andriotis2019ManagingLearning}. As DDMAC adjusts the weights of the actor networks according to noisy rewards collected at the system level, DDMAC policies intrinsically consider system-effects stemming from structural and statistical dependencies. Through numerical experiments, we demonstrate the efficacy of the proposed method for I\&M planning of structural systems exposed to fatigue deterioration. In particular, the effects of including deterioration dependence and campaign cost models are explored for the case of a 9-out-of-10 system. In the second application studied, featuring a steel frame structural system, the focus is on examining and interpreting the inherent allocation of maintenance interventions by DDMAC policies according to the element importance to the global structural reliability. In all experiments analyzed, DDMAC policies are compared against state-of-the-art optimized heuristic policies.

The remainder of the paper is structured as follows: an overview of POMDP methods along with the proposed factored formulation are presented in Section \ref{Sec:factPOMDP}. In Section \ref{Sec:sysEff}, the definition and modeling of Gaussian hierarchical structures are introduced, together with a belief update algorithm, applicable to environments under general deterioration dependence. The integration of the simulator, defined as a factored POMDP, with DDMAC is presented in Section \ref{Sec:DRL}. The numerical experiments are then introduced and discussed in Section \ref{Sec:experim}, concluding with some final remarks in Section \ref{Sec:conclusions}.

\section{I\&M decision problem formulated as a factored POMDP}\label{Sec:factPOMDP}
\subsection{Factored POMDP definition}
The inspection and maintenance (I\&M) planning decision-making problem is formulated here as a Partially Observable Markov Decision Process (POMDP), whose transition and observation models are defined by Bayesian network structures. POMDPs provide a principled mathematical framework for optimal planning and decision-making under uncertainty, formally specified by the tuple $\langle \mathcal{S},\mathcal{A},\mathcal{O},\mathcal{T},\mathcal{Z},\mathcal{R},\gamma \rangle$. A decision maker (henceforth agent) interacts with a stochastic environment, described by the state $s\in \mathcal{S}$, taking actions $a \in \mathcal{A}$ over a finite or infinite horizon $t_N$. The dynamics correspond to those in a MDP: At each time step $t$, an agent takes an action $a_t \in \mathcal{A}$, and the environment evolves from state $s_t \in \mathcal{S}$ to state $s_{t+1} \in \mathcal{S}$, according to the transition model $\mathcal{T} \coloneqq p(s_{t+1}|s_t,a_t)$. In a MDP, the agent receives a reward based on the cost model $\mathcal{R} \coloneqq r_t(s_t,a_t,s_{t+1})$ discounted by factor $\gamma$, and the objective is to find the policy $\pi^*$ that induces the optimal value function $V^*(s_t)$:
\begin{equation}\label{Eq:MDPvalue}
V^*(s_t) = \max_{a_t\in \mathcal{A}} \left\{ r(s_t,a_t) + \gamma \sum_{s_{t+1}\in \, \mathcal{S}} p(s_{t+1}|s_t,a_t)V^*(s_{t+1}) \right\}
\end{equation}
In a POMDP, however, states $s \in \mathcal{S}$ are not directly observed and, instead, observations $o \in \mathcal{O}$ can be collected according to the observation model $\mathcal{Z} \coloneqq p(o_{t+1}|s_{t+1},a_t)$. Note that the observation model is the likelihood of collecting an observation $o_{t+1}\in O$ after taking an action $a_t$ and having transitioned to state $s_{t+1}$. In an I\&M context, the observation model is often modeled by Probability of Detection (PoD) curves or according to inspection/monitoring measurement noise \cite{morato2022optimal}. A POMDP policy is a mapping of the dynamically updated history of actions and observations, $a_{0:t-1}, o_{0:t}$, to the current action $a_t$. This history is sufficiently encoded in belief $\mathbf{b}$, which is the probability over system states, b(s). The optimal policy $\pi^*$, therefore, corresponds to the value function \cite{sarsop_POMDP} that satisfies the Bellman equation:      
\begin{equation}\label{Eq:POMDPvalue}
V^*(\mathbf{b}_t) = \max_{a_t\in A} \left\{ \sum_{s_t\in \mathcal{S}}r(s_t,a_t)b(s_t) + \gamma \sum_{o_{t+1}\in \mathcal{O}} p(o_{t+1}|\mathbf{b}_t,a_t)V^*(\mathbf{b}_{t+1}) \right\},
\end{equation}
where $p(o_{t+1}|\mathbf{b}_t,a_t)$ is the probability of collecting an observation $o_{t+1}\in O$ given the belief $\mathbf{b}_t$ and action $a_t\in A$. Assuming a Markovian environment is reasonable in most practical applications with the aid of state augmentation techniques \citep{Papakonstantinou2014Part1}, hence, any general I\&M planning decision problem can be efficiently formulated as a POMDP. The determination of the optimal I\&M policy $\pi^*$ becomes the main objective, inducing a minimization of the expected life-cycle costs $r_{tot}$, by balancing structural failure risk against inspection and maintenance costs:
\begin{equation}\label{Eq:IMCosts}
\mathbb{E}[r_{tot}] = \sum_{t=0}^{t_N} \big[ \gamma^t (r_{ins,t} + r_{rep,t} + r_{F,t}) \big],
\end{equation}
where $r_{ins}$, $r_{rep}$ and $r_{F}$ stand for inspection, repair and failure costs, respectively, defined as negative rewards. In terms of utilities, the failure risk $ r_{F}$ is typically defined in a structural reliability context as the annual probability of a failure event weighted by the consequence of a structural failure, which might also include environmental and societal consequences, specified in equivalent units. The definition of the failure risk at the system level will be further elaborated in Section \ref{Sec:sysEff}.  

Existing I\&M planning applications often model the deterioration evolution $d$, at the component level, conditional on a set of random variables $\boldsymbol{\uptheta_d}$ \cite{StraubDBN2009,LuqueDBN2019,DBNLuque2016} or as a function of the deterioration rate $\tau$ \cite{Papakonstantinou2014Part2,Andriotis2019ManagingLearning}. Both formulations are equivalent for modeling deterioration processes, as already discussed and demonstrated in \cite{morato2022optimal} and shown in Fig. 1. When observations are collected, through inspections or monitoring, Bayesian updating can be then conducted. Available algorithms allow exact Bayesian inference if the problem is formulated in a discrete state space \cite{murphythesis}, as the computation of Bayes' normalization constant is a challenging task in continuous state spaces \cite{papakonstantinou2022scaled}. In order to utilize discrete state based algorithms, the involved continuous random variables can be discretized. The quality of the discretization has a huge impact and shall be treated carefully \cite{StraubDBN2009,morato2022optimal}, especially when the problem deals with rare events, e.g., failure events. In general, an efficient discretization aims at minimizing the computational expense while preserving the required level of accuracy. 

In a POMDP, the states cannot be directly observed and the decision maker reasons under partial observability, only informed by a belief $\mathbf{b}$, which is defined as the probability over states. At each time step, the belief is dynamically updated, based on Bayes' rule, depending on the initial belief, $\mathbf{b}_t$, the action taken, $a_t$, and the collected observation, $o_t$, following three main steps: (i) the belief evolves according to the transition model $p(s_{t+1}|s_t,a_t)$, (ii) the belief is updated based on the collected observation with probability $p(o_{t+1}|s_{t+1},a_t)$, and (iii) the belief state is normalized. This belief update operation is denoted as forward pass within the context of hidden Markov models \cite{murphythesis}. At the system level, the belief of each component can be updated by implementing the steps listed in Algorithm \ref{alg:beliefUpd}. 

\begin{algorithm}
\caption{Belief update for a system of $N_c$ components}
\begin{algorithmic}

\Function{updateBelief}{$\mathbf{b}_t,{a}_t,{o}_{t+1}$}       
	\For{$1,N_c$}
		\State $b(s_{t+1}) \leftarrow b(s_t)\,p(s_{t+1}|s_t,a_t) $ \Comment{propagation step}
		\State $b(s_{t+1}) \leftarrow b(s_{t+1})\,p(o_{t+1}|s_{t+1},a_t) $ \Comment{estimation step}
		\State $b(s_{t+1}) \leftarrow b(s_{t+1})/  p(o_{t+1}|\mathbf{b_t},a_t)$ \Comment{normalization step} 
	\EndFor    
\EndFunction

\end{algorithmic}
\label{alg:beliefUpd}
\end{algorithm}

State-of-the-art POMDP solvers often require the modeling of the POMDPs in a flat structure, which can be usually encoded by augmenting the state space \cite{Papakonstantinou2014Part1}, particularly if the process is described by multiple random variables. However, POMDPs can also be formulated in a factored fashion, exploiting the dependence structure among random variables and thus significantly alleviating the required computational effort. We specify here the transition and observation models based on conditional structures described by dynamic Bayesian networks (DBNs), and while the belief state $\mathbf{b}$ remains the same as that for flat POMDPs, the transition and observation models are now constructed by taking advantage of the involved dependencies. For instance, the deterioration rate model can be constructed as $p(d_{t+1}|d_t,\tau_{t+1})\,p(\tau_{t+1}|\tau_{t}) $ instead of $p(d_{t+1},\tau_{t+1}|d_{t},\tau_{t})$. This incorporation of conditional structures allows a reduction of the transition model dimensionality from $|\mathcal{S}_d|^2|\mathcal{S}_\tau| ^2$ to  $|\mathcal{S}_d|^2|\mathcal{S}_\tau| + |\mathcal{S}_\tau| ^2$ and can achieve significant computational benefits when multiple random variables are involved. This formulation can be seamlessly applied to simulate the deterioration environment, as will be explained in Section \ref{Sec:DRL}, due to the flexibility naturally offered by the proposed deep reinforcement learning approach.  

\section{System effects in I\&M planning}\label{Sec:sysEff}
\subsection{Deterioration dependence in a hierarchical Gaussian structure}
\begin{figure}[h]
	\centering
		\includegraphics[scale=1]{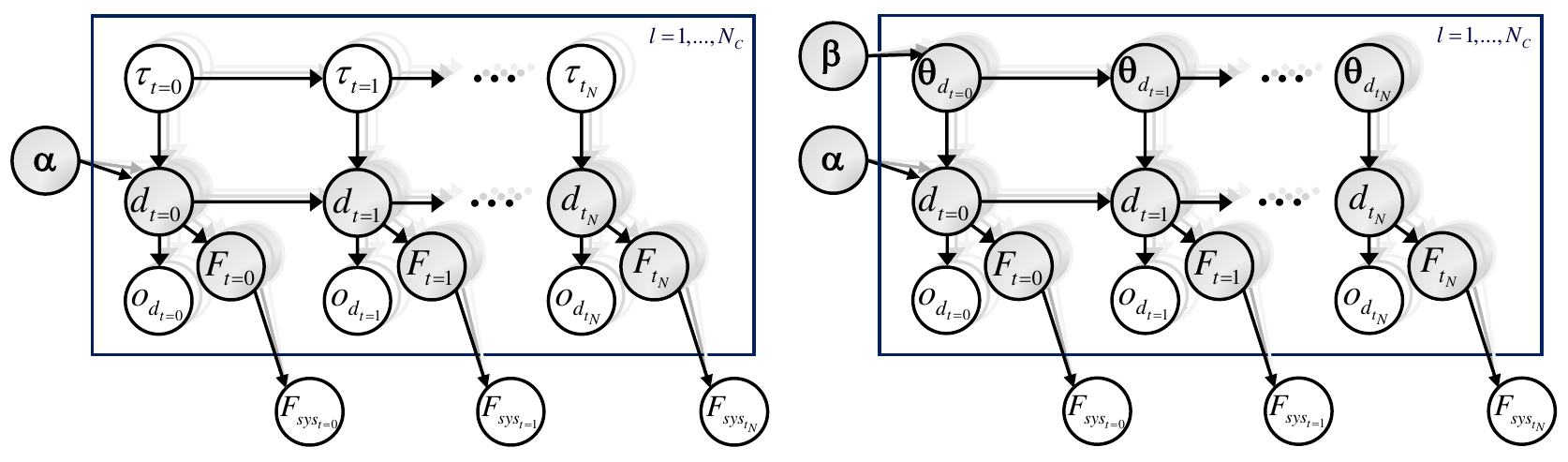}
	\caption{Graphical representation of dynamic Bayesian networks for modeling deterioration processes. At the component level, the damage $d_t$ evolves over time $t$ as a function of the deterioration rate $\tau_t$ (left) or conditional on a set of parameters $\boldsymbol{\uptheta_{d_t}}$ (right). While $d_t$ and $\boldsymbol{\uptheta_{d_t}}$ are hidden states, partially observed through $o_{d_t}$, $\tau_t$ is fully observable. The deterioration dependence among components is encoded by the hyperparameter or set of hyperparameters $\boldsymbol{\upalpha}, \boldsymbol{\upbeta}$. The binary failure observable state indicates either the survival or failure state of the system $F_{sys_t}$ depending on the components failure hidden state $F_t$. Note that other variants of the system failure formulation can also be represented accordingly.}
	\label{FIG:dbnDr}
\end{figure}
Existing methods model the deterioration correlation among components either via random fields or through common influencing factors. Whereas the former are particularly useful for applications in which the dependence is attributed to the geometrical distance between components, the latter are more suitable for systems in which identical attributes of physical phenomena, e.g., similar manufacturing techniques or similar loading, lead to shared sources of model uncertainties among the components \cite{DBNLuque2016}. In a hierarchical structure, the deterioration of each component is defined conditional on a set of common influencing variables, shared among all the components and represented at the highest level of the hierarchy. In theory, the state space of a system under deterioration dependence can be modeled directly as the joint space of all the parameters involved in the deterioration process of the system. In this case, the discretized state space would grow exponentially with the number of components $N_C$ included, into a $|S|^{N_C} $ dimensional space. To overcome this increase in dimensionality, we adopt the hierarchical Gaussian structure previously proposed in \cite{DBNLuque2016}, in which the belief state of each component is encoded conditional on a hyperparameter (common influencing variable) $\alpha$ or set of hyperparameters $\boldsymbol{\alpha}$. The central idea behind this hierarchical structure is that component beliefs for a given hyperparameter $\mathbf{b}(\mathbf{s}|\alpha)$ are independent, enabling an efficient decoupling of the components joint space. This decoupling alleviates the computational complexity from the original joint space $|\mathcal{S}|^{N_C} $ to a space $|\mathcal{S}|\cdot |\mathcal{S}_{\alpha}|\cdot N_C$ that grows linearly with the number of considered components $N_C$. Note that the state space includes now the states of the hyperparameter(s), which should also be properly discretized. The increase of the state space due to the incorporation of the hyperparameter(s) is however less significant than when considering the joint state space. 

A graphical representation of the proposed hierarchical structure is illustrated in Fig. \ref{FIG:dbnDr}, applicable to deterioration processes modeled either as a function of the deterioration rate or conditional on a set of parameters \cite{morato2022optimal}. In either case, the deterioration process $d$ is encoded conditional on the hyperparameter(s) $\boldsymbol{\upalpha}$, along with the deterioration rate $\tau$ or parameters $\boldsymbol{\uptheta}_d$. Evidence collected through observations $\mathbf{o}_{d_t}$ does not only serve for updating the damage state, but also for updating the hyperparameters. Since the hyperparameters are parent nodes for all the components, once a component is inspected, the hyperparameters are also updated, influencing all the other components, even those for which evidence was not directly available. The reliability of the system is represented in Fig. \ref{FIG:dbnDr} by the binary node $F_{sys}$, conditional on the failure state of the components $F^{(l)}$. At a component level, the failure state is modeled by the binary variable $F^{(l)}$ and corresponds to the subset of the deterioration space classified as failure $\mathcal{S}_F \subseteq \mathcal{S}$.  

This Gaussian hierarchical structure is a mathematically motivated model induced by the convenient formulation available for normal random variables, i.e., the conditional and joint distributions of normal random variables are also normally distributed. Let us first consider the special case in which the marginal probability of each considered component deterioration is defined as a standard normal random variable $Y_i$. Under correlation, the parameters $Y_i$ are, however, defined as normal random variables with mean $\lambda_i \alpha$ and standard deviation $\sqrt{1-\lambda_i^2} $\cite{dunnett1955approximations}:
\begin{equation}\label{Eq:hypDef1}
Y_i= \sqrt{1-\lambda_i^2}X_i + \lambda_i \alpha
\end{equation}
Since both $X_i$ and $\alpha$ are independent standard normal random variables, the covariance of $Y_i$ and $Y_j$ can be formulated as:
\begin{equation}\label{Eq:hypCov}
cov(Y_i,Y_j)= (1-\lambda_i^2)cov(X_i,X_j) + \sqrt{1-\lambda_i^2}(\lambda_j)cov(X_i,\alpha) +  \sqrt{1-\lambda_j^2}(\lambda_i)cov(X_j,\alpha)+ \lambda_i\lambda_j cov(\alpha,\alpha)
\end{equation}     
After removing all the terms associated with zero covariance, i.e., $cov(X_i,X_j)$, $cov(X_i,\alpha)$, and $cov(X_j,\alpha)$, we can define the correlation coefficient between $Y_i$ and $Y_j$ as:
\begin{equation}\label{Eq:hypCorr}
\rho(Y_i,Y_j)= \lambda_i\lambda_j
\end{equation}  
If all the components are equi-correlated, then $\lambda_i=\lambda_j=\sqrt{\rho(Y_i,Y_j)}$, for all $i,j$. This presented Gaussian structure is further generalized in this work for the general case of unequally correlated components, by preserving the validity of Eq. \ref{Eq:hypCorr}. For complex correlation configurations, one hyperparameter $\alpha$ might not be sufficient to satisfy Eq. \ref{Eq:hypCorr}, and in that case, one can incorporate additional hyperparameters $\boldsymbol{\alpha}$, at the expense of a higher computational cost. When multiple hyperparameters $\boldsymbol{\alpha}$ are included, the best fit for $\rho(Y_i,Y_j)= \sum_{k=1}^{m}(\lambda_{ik}\lambda_{jk})$ can be found via optimization procedures, e.g., least squares \cite{song2009system}. 
Once the Gaussian correlation structure is specified through the parameters $\boldsymbol{\lambda}$, the cumulative distribution of $Y_i$ conditional on the hyperparameter(s) $\boldsymbol{\alpha}$ can be defined as:
\begin{equation}\label{Eq:belGauss}
F_{Y_i|\alpha}(y_i)=\Phi\left[\frac{y_i-\lambda_i \alpha}{\sqrt{1-\lambda_i^2}}\right]
\end{equation}
For the cases in which the deterioration process is modeled by random variables other than Gaussian and considering that a Nataf transformation is applicable \cite{DBNLuque2016}, then Eq. \ref{Eq:belGauss} can be also redefined as:
\begin{equation}\label{Eq:belRvs}
F_{D_i|\alpha}(d_i)=\Phi \left[\frac{ \Phi^{-1}[F_d(d_i)] -\lambda_i \alpha}{\sqrt{1-\lambda_i^2}}\right]
\end{equation}
where $F_{D_i|\alpha}(d_i)$ stands for the cumulative distribution function of a variable $d_i$ conditional on the hyperparameter(s) $\boldsymbol{\upalpha}$, and $\Phi$ is the standard normal cumulative distribution function. In a discrete state space, the belief conditional on the hyperparameters is equal to the difference between the cumulative distribution function at the upper boundary and at the lower boundary of each belief interval:
\begin{equation}\label{Eq:belCond}
b(d_i|\alpha)= F_{D_i|\alpha}(d_i^{+}) - F_{D_i|\alpha}(d_i^{-})
\end{equation}
\subsection{Belief update under deterioration dependence}
We reformulate here the belief update algorithmic scheme introduced in Section \ref{Sec:factPOMDP} for a system under deterioration dependence among components. All necessary implementation steps are listed in Algorithm \ref{alg:beliefUpdCorr}. Bayesian inference is firstly conducted for the conditional beliefs ${b}(s_{t+1}|\alpha)$ and hyperparameters ${b}({\alpha})$, propagating uncertainty according to the transition model $p(s_{t+1}|s_t,a_t)$ and observation model $p(o_{t+1}|s_{t+1},a_t)$. The likelihood of collecting an observation given the hyperparameter(s) $p(o_{t+1}|\alpha)$, later necessary to update ${b}(\alpha)$, can be easily computed by marginalizing out the states other than $\alpha$:  
\begin{equation}\label{Eq:margHyp}
p(o_{t+1}|\alpha,a_t)=\sum_{s_{t+1}\in \mathcal{S}}\Big[b(s_{t+1}|\alpha)\,p(o_{t+1}|s_{t+1},a_t)\Big]
\end{equation}
Bayesian inference is then conducted for the hyperparameters:
\begin{equation}\label{Eq:infHyp}
p(\alpha|o_{t+1},a_t)=b(\alpha)p(o_{t+1}|\alpha,a_t) / p(o_{t+1}|a_t)
\end{equation}
After updating the conditional beliefs and common influencing variables, the marginal deterioration beliefs can be computed by marginalizing out the hyperparameters $\alpha$ as:
\begin{equation}\label{Eq:margBel}
b(s_{t+1}) = \sum_{\alpha \in \Gamma} \Big[p(s_{t+1}|\alpha)\, b({\alpha}) \Big]
\end{equation}
\begin{algorithm}
\caption{Belief update under deterioration dependence for a system of $N_C$ components}
\begin{algorithmic}

\Function{updateBelief}{${b}(s_t|\alpha), {b}(\alpha), {a}_{t},{o}_{t+1} $}       
	\For{$1,N_c$}
		\State $b(s_{t+1}|\alpha) \leftarrow b(s_{t}|\alpha)\,p(s_{t+1}|s_t,a_t) $ \Comment{propagation step}
		\State $b(s_{t+1}|\alpha) \leftarrow b(s_{t+1}|\alpha)\,p(o_{t+1}|s_{t+1},a_t) $ \Comment{estimation step}
		\State $b(s_{t+1}|\alpha) \leftarrow b(s_{t+1}|\alpha)/ p(o_{t+1}|\mathbf{b},a_t) $ \Comment{normalization step}
		\State $p(o_{t+1}|\alpha) \leftarrow \sum_{s_{t+1}\in S} [b(s_{t+1}|\alpha)\,p(o_{t+1}|s_{t+1},a_t)]$ \Comment{likelihood}
		\State $b({\alpha}) \leftarrow b({\alpha})\,p(o_{t+1}|\alpha,a_t)/ p(o_{t+1}|a_t)  $ \Comment{hyperparameter(s) update}
	\EndFor    
	\For{$1,N_c$}
		\State $b(s_{t+1}) \leftarrow \sum_{\alpha \in \boldsymbol{\alpha}} [b(s_{t+1}|\alpha)\, b({\alpha}) ]$ \Comment{marginalizing out hyperparameter(s)}
	\EndFor
	\State \Return $\mathbf{b}(s_{t+1})$    
\EndFunction

\end{algorithmic}
\label{alg:beliefUpdCorr}
\end{algorithm}

The effect of maintenance actions on the Gaussian dependence structure has not been explored in the existing literature \cite{LuqueDBN2019,DBNLuque2016}, up to the best knowledge of the authors. Whereas the defined deterioration dependence is preserved if no maintenance interventions are planned, structural interventions can potentially disrupt the underlying correlation structure. For instance, if a structural system is specified with a correlated initial crack size among fatigue hotspots, this correlation structure will be perturbed after a component is repaired, along with the correlation reduction naturally experienced by the system over time. The correlation evolution associated with the latter is intrinsically quantified through the uncertainty propagation and updating operations formulated in Eqs. \ref{Eq:margHyp}-\ref{Eq:infHyp}, whereas the correlation disruption associated with the former can be modeled by now defining the transition model $p(s_{t+1}|s_t,a_t, \boldsymbol{\alpha})$ of the involved components also conditional to the hyperparameter(s) $\boldsymbol{\alpha}$, enabling therefore the removal or modification of the deterioration dependence by redefining and implementing the relevant correlation coefficients $\lambda_i$ in Eqs. \ref{Eq:belGauss} - \ref{Eq:belCond}. Additional discussion and implementation of this aspect is presented in the numerical experiments section.

\subsection{System structural reliability and system cost model}
As input to the I\&M decision-making problem (Section \ref{Sec:factPOMDP}), utilities, $r_{ins}$ and $r_{rep}$, are assigned to inspection and repair actions, respectively, specified according to available options and settings in each problem. The annual risk of a system failure, $r_F$, is defined accounting for two consecutive time steps (e.g., years) and its associated system failure cost, $r_f$, as: 
\begin{equation}\label{Eq:pfSys1}
r_F = (p_{F_{sys,t+1}} - p_{F_{sys,t}}) r_f 
\end{equation}
The system structural failure event, as illustrated in Figure \ref{FIG:dbnDr} by the node $F_{sys}$, is specified by a binary variable $p_{F_{sys}}$, indicating the failure and survival states, conditional on the belief state $b(s)$ of the structural components, as these are determined by the performed I\&M actions in time. In principle, $p_{F_{sys}}$ could be directly defined as a function of the components belief state; in practice, however, $p_{F_{sys}}$ remains only conditional to the event of component failures $p_F$, specified as:
\begin{equation}\label{Eq:failComp}
p_F = \sum_{s \in \mathcal{S}_F} b(s)
\end{equation}
where $\mathcal{S}_F$ corresponds to the components state subset classified as failure $\mathcal{S}_F \subseteq \mathcal{S}$. Within the deep reinforcement learning approach presented in Section \ref{Sec:DRL}, $P_{F_{sys}}$ can be computed via closed-form  procedures and/or supported by efficient matricial algorithms \cite{song2009system}; or it can be computed following a general scheme, obtaining $p_{F_{sys}}$ through a simulator \cite{derkiureghian_2022}. By assigning utilities to the system state, the importance of each structural element to the global risk of a system failure is implicitly accounted. To illustrate the effect of defining the failure risks at the system level, I\&M strategies for a redundant 2-dimensional frame structure are later explored in Section \ref{Sec:experim}.

In most structural systems, from bridges to offshore platforms or wind farms, inspection and repair actions are not planned separately for each structural element. Maintenance campaigns are instead scheduled, collecting information or performing repairs on a group of structural components. The cost model can thus be adapted from Eq. \ref{Eq:IMCosts}, to include a fixed campaign cost, $r_{camp}$, incurred every time a campaign is planned, along with inspection, $r_{ins}^{(l)}$, and repair, $r_{rep}^{(l)}$, costs assigned to the individual components according to any nonlinear function of choice, $\mathcal{H}(.)$, e.g., simple linear sum operator, as:
\begin{equation}\label{Eq:costCamp}
r_{tot}= r_{camp} + \mathcal{H} \left( r_{ins}^{(l)}, r_{rep}^{(l)} \right) + r_F
\end{equation}

\section{Optimal I\&M planning via deep reinforcement learning}\label{Sec:DRL}
I\&M planning decision problems, formulated as POMDPs (as explained in Sections \ref{Sec:factPOMDP} and \ref{Sec:sysEff}), can be solved by dynamic programming algorithms, e.g., via exact alpha-vector value iteration \cite{Piineau_POMDP_2003}. In practice, however, exact value iteration can be applied to only very small state space problems due to the complexity associated with the exponential increase in the number alpha vectors with the number of observations at every iteration. Recently, I\&M planning decision problems, at the component level, formulated as POMDPs and characterized by multiple states have been efficiently solved via point-based POMDP algorithms  \cite{morato2022optimal,Papakonstantinou2014Part2,Kostas_MOMDP_POMDP}. Point-based solvers exploit the fact that the value function $V(\mathbf{b})$ (Eq. \ref{Eq:POMDPvalue}) is piece-wise linear and convex and can be thus parameterized by a set of $\mathbf{v}_p\in \mathcal{V}_p$ vectors, each of which is associated with a specific action $a\in \mathcal{A}$. The optimal value function, $V^{*}(\mathbf{b})$, can be therefore defined in terms of a set of $\mathbf{v}_p$ vectors \cite{puterman2014markov}: 
\begin{equation}\label{Eq:polPoint}
V^*(\mathbf{b})={max_{\mathbf{v}_p\in \mathcal{V}_p}}\left[\sum_{s\in \mathcal{S}} b(s) v_p(s)\right]
\end{equation}
State-of-the-art point-based POMDP solvers mainly differ on their approach of sampling reachable belief points, and the way Bellman backup operations are executed, e.g., in \cite{sarsop_POMDP, PerseusCONFSpaan2005, FRTDP_Conf2006}. The reader is directed to \cite{Kostas_MOMDP_POMDP} for a detailed comparison of point-based solvers applied to infrastructure I\&M settings. While point-based solvers are able to efficiently provide optimal policies at the component level and for realistically large systems, the dimensionality still becomes a limiting factor in high-dimensional state, action, and observation space settings, typical in structural systems. Deep Reinforcement Learning (DRL) provides then a powerful solution in such settings, as the value or policy function can be parameterized with deep artificial neural networks. Thereby, the planning task reduces to finding a number of parameters that is much lower than the number of original states and actions of the problem. The interested reader is directed to \cite{sutton2018reinforcement, li2017deep} for a well elaborated introduction and discussion on DRL. In our proposed approach, we integrate the factored POMDP formulation introduced in Section \ref{Sec:factPOMDP} with a Decentralized Deep Multi-agent Actor-Critic (DDMAC) scheme, adopted from \cite{Andriotis2019ManagingLearning}. This combination provides an efficient algorithmic platform for inspection and maintenance planning of structural systems under deterioration, reliability and cost dependencies, in large-scale multi-component environments. 

\begin{figure}
	\centering
		\includegraphics[scale=1]{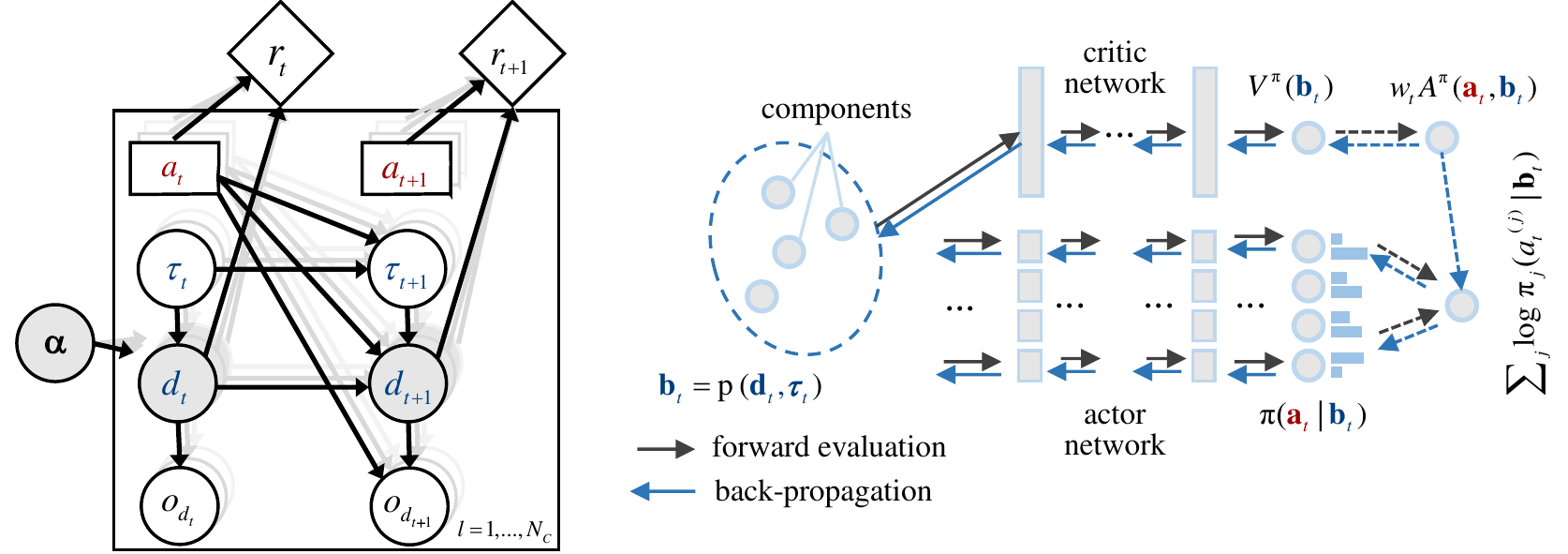}
	\caption{On the left: Representation of a factored POMDP derived from the deterioration rate dynamic Bayesian network introduced in Fig. \ref{FIG:dbnDr}. The deterioration process $d_t$, influenced by the deterioration rate $\tau_t$, conditional on the hyperparameters $\boldsymbol{\alpha}$ and partially observed through $o_{d_t}$, is controlled by the action decision node $a_t$. A reward $r_t$ is collected as a result of taking action $a_t$ at state $d_t$. On the right: Deep Decentralized Multi-Agent Actor Critic (DDMAC) featuring the critic network at the top, and a group of actor networks, one for each component, at the bottom.}
	\label{FIG:HDDMAC}
\end{figure}

Each component of the system is controlled by the stochastic policy $\pi(a|\mathbf{b},\boldsymbol{\uptheta}^{\pi})$ provided by a group of multi-agent actor networks, defined as a function of parameters $\boldsymbol{\uptheta}^{\pi}$, as illustrated on the right side of Fig. \ref{FIG:HDDMAC} with light blue bars. In many applications, DRL policies after training are nearly deterministic, suggesting one action in particular, whereas stochastic policies are often optimal in constraint environments \cite{andriotis2021deep}. In our implementation, we consider agents acting as independent units in a decentralized manner, i.e., the actions taken by one actor are naturally not affected by the actions taken by other actors:
\begin{equation}\label{Eq:actIndep}
\pi(\mathbf{a}| \mathbf{b}) = \prod_{l=1}^{N_C} \pi_l (a^{(l)}|\mathbf{b})
\end{equation}

The input to the actor networks corresponds to the marginal belief states of all components along with the deterioration rate and the time step encoded as a zero-one vector. For instance, if the environment is described by the factored POMDP represented on the left side of Fig. \ref{FIG:HDDMAC}, the actor networks receive the deterioration belief states $\mathbf{b}(s_d)$ and deterioration rate states $\mathbf{b}(s_{\tau})$ for all components  (in $|\mathcal{S}_d|\cdot N_c$ and $|\mathcal{S}_{\tau}|\cdot N_c$ matrix formats, respectively), plus an input indicating the time step $t \in t_N$. If deterioration dependence is included through a hierarchical Gaussian model, as explained in Section \ref{Sec:sysEff}, then conditional beliefs $\mathbf{b}(s_d|\boldsymbol{\upalpha})$ and hyperparameters beliefs $\mathbf{b}(\boldsymbol{\upalpha})$ should also be used while simulating the deterioration environment. Even for environments under deterioration dependence, the neural networks only receive as input the components' marginal beliefs $\mathbf{b}^{(l)}$, for all $l$, computed by following the steps listed in Algorithm \ref{alg:beliefUpdCorr}. ReLU activation functions are used for the hidden layers of the actor networks, and the output layer is activated by a softmax function, generating the output policy as a probability distribution over the available actions.    

The actor network weights are adjusted/updated according to the noisy rewards collected from a batch of previous experiences, following an off-policy training approach that offers more sample efficiency than on-policy training algorithms. A replay buffer \cite{schaul2015prioritized} stores beliefs $\mathbf{b}_t$, actions $\mathbf{a}_t$, rewards $r(\mathbf{b}_t,\mathbf{a}_t)$ and behavior policies $\mu_t$, experienced during the simulations of the deterioration environment. The off-policy gradient estimator is thus formulated with samples generated by a behavior policy $\mu$, different from $\pi$, and corrected with the truncated importance sampling weight $w_t=\text{min}\{c,\pi(\mathbf{a}_t|\mathbf{b}_t)/\mu(\mathbf{a}_t|\mathbf{b}_t)\}$, with $c>0$ \cite{Andriotis2019ManagingLearning}:
\begin{equation}\label{Eq:gradActors}
\mathbf{g}_{\boldsymbol{\uptheta}^{\pi}} = \mathbb{E}_{\,\mathbf{a}_t\sim \boldsymbol{\upmu}} \Bigg[  w_t \Bigg\{\sum_{i=1}^{N_c} \nabla_{\boldsymbol{\uptheta}^{\pi}} \, log \, \pi_i (a_t^{(i)}|\mathbf{b}_{t}, \boldsymbol{\uptheta}^{\pi})\Bigg\}A^{\pi}(\mathbf{b}_t,\mathbf{a}_t|\boldsymbol{\uptheta}^V) \Bigg]
\end{equation}
The advantage function $A^{\pi}(\mathbf{b}_t,\mathbf{a}_t)$ indicates how optimal is action $\mathbf{a}_t$ with respect to the current estimated value function $V^{\pi}(\mathbf{b}_{t})$ and defined in a temporal difference learning fashion as:
\begin{equation}\label{Eq:advanFunc}
A^{\pi}(\mathbf{b}_t,\mathbf{a}_t|\boldsymbol{\uptheta}^V) \simeq r(\mathbf{b}_t,\mathbf{a}_t)+ \gamma V(\mathbf{b}_{t+1}| \boldsymbol{\uptheta}^{V}) - V(\mathbf{b}_t| \boldsymbol{\uptheta}^{V})
\end{equation}
The value function is approximated by the critic network, defined as a function of parameters $\boldsymbol{\uptheta}^{V}$, as illustrated on the right side of Fig. \ref{FIG:HDDMAC}. Whereas the critic network receives the same input as the actor network (components marginalized beliefs, deterioration rates, and a time step indicator), the output of the critic is the value function, i.e., one scalar value that indicates the expected reward of the system. The critic network provides the value function used by the advantage function $A^{\pi}(\mathbf{b}_t,\mathbf{a}_t|\boldsymbol{\uptheta}^V)$, acting, therefore, as a critic who is determining how good the action taken by the actor network is. The training of the critic network also follows a temporal difference approach, collecting experiences from the replay buffer, and adjusting the critic parameters $\boldsymbol{\uptheta^{V}}$ according to the gradient:
\begin{equation}\label{Eq:gradCritic}
\mathbf{g}_{\boldsymbol{\uptheta}^{V}} = \mathbb{E}_{\,\mathbf{a}_t\sim \boldsymbol{\upmu}}  \Big[ w_t \nabla_{\boldsymbol{\uptheta}^V}V^{\pi}(\mathbf{b}_{t}|\boldsymbol{\uptheta}^{V})A^{\pi}(\mathbf{b}_t,\mathbf{a}_t|\boldsymbol{\uptheta}^V)\Big] 
\end{equation}

All the algorithmic steps are described in Algorithm \ref{alg:DDMAC}. With our proposed method, we are able to find optimal I\&M policies for structural systems featuring very high dimensional state, action and observation spaces. Moreover, the obtained DDMAC policies intrinsically account for system-effects (Section \ref{Sec:sysEff}) as the actor network is adjusted according to the rewards collected by simulating the deteriorating environment at the system level. Specifically, the integration of DDMAC with a deterioration environment simulated with a factored POMDP (Section \ref{Sec:factPOMDP}) enables the identification of optimal I\&M policies that concurrently consider the following system effects:  
\begin{enumerate}[label=\roman*)]
\item Deterioration dependence among components (statistical dependence): A Gaussian hierarchical model efficiently captures the deterioration dependence, e.g., initial crack size, or loading. The belief of each component is conditional on the common hyperparameter(s) $\mathbf{b}(s_d|\boldsymbol{\alpha})$. Under statistical deterioration dependence, information collected by inspecting one component informs belief updates for other components as per the specified deterioration. The influence of this system-effect on the policy is explored via numerical experiments in Section \ref{Sec:experim}, for a 9-out-of-10 system and for a steel frame structural system subject to fatigue deterioration.   
\item System structural reliability (structural dependence): Failure risk is computed at the system level, by multiplying annual risk with a negative reward $r_f$ that is defined as a function of the components structural health, as shown in Eq. \ref{Eq:pfSys1}. The actors, even though acting individually, are all conditioned on all component beliefs, thus knowing the system reliability. DDMAC is able to intrinsically adjust the policy according to the relative importance of each component to the system structural reliability, as demonstrated with the numerical experiments conducted for the steel frame structural system (Section \ref{Sec:experim}).  
\item Inspection and maintenance cost model (cost dependence): A campaign cost $r_{camp}$ is included, in the applications, as a base cost, if at least one component is inspected or repaired, plus an additional inspection or repair cost for each inspected or repaired component, as shown in Eq. \ref{Eq:costCamp}. Since DDMAC collects rewards at the system level, the campaign cost model affects the resulting I\&M policies, concentrating inspection and repair actions at particular time steps, as observed in the numerical experiments conducted for the 9-out-of-10 system (Section \ref{Sec:experim}).  
\end{enumerate}
\begin{algorithm}
\caption{Deep Decentralized Multi-agent Actor Critic (DDMAC)}
\begin{algorithmic}

    \State Initialize replay buffer
    \State Initialize actor and critic network weights $\theta^{\pi}, \theta^{V}$  
	\For{$episode=1,M$}
		\For{$t=1,t_N$}
		\State Select action $\mathbf{a}_t$ at random according to exploration noise 
		\State Otherwise select action $\mathbf{a}_t \sim \boldsymbol{\upmu}_t = \{\pi_j(\cdot|\mathbf{b}_t,\theta^{\pi} ) \}^{N_c}_{j=1} $ 
		\State Collect reward $r(\mathbf{b}_t,\mathbf{a}_t)$ 
		\State Observe $o_{t+1}^{(l)} \sim p(o_{t+1}^{(l)}|\mathbf{b}_t,\mathbf{a}_t)$ for $l=1,2,...,N_c$
		\State Compute beliefs $\mathbf{b}_{t+1}$: \Call{updateBelief}{$\mathbf{b}_t, \mathbf{a}_t,\mathbf{o}_t$}
		\State Store experience $(\mathbf{b}_t, \mathbf{a}_t, \mu_t, r(\mathbf{b}_t,\mathbf{a}_t),  \mathbf{b}_{t+1})$ in replay buffer
		\State Sample batch of $(\mathbf{b}_i, \mathbf{a}_i, \mu_i, r(\mathbf{b}_i,\mathbf{a}_i),  \mathbf{b}_{i+1})$ from replay buffer
		\State If $\mathbf{b}_{i+1}$ is terminal state $A_i^{\pi}=r(\mathbf{b}_i,\mathbf{a}_i) - V^{\pi}(\mathbf{b}_i, \theta^{V})$
		\State Otherwise $A_i^{\pi}=r(\mathbf{b}_i,\mathbf{a}_i)+ \gamma V^{\pi}(\mathbf{b}_{i+1}, \theta^{V}) - V^{\pi}(\mathbf{b}_i, \theta^{V})$
		\State Update actor parameters $\theta^{\pi}$ according to gradient:
		\State $\;\;\; \mathbf{g}_{\boldsymbol{\uptheta}^{\pi}} \simeq  \sum_i w_i \{\sum_{j=1}^{N_c} \nabla_{\boldsymbol{\uptheta}^{\pi}} \, log \, \pi_j (a_i^{(j)}|\mathbf{b}_{i}, \theta^{\pi})\}A_i^{\pi}$
		\State Update critic parameters $\boldsymbol{\uptheta}^{V}$ according to gradient:
		\State $\;\;\; \mathbf{g}_{\boldsymbol{\uptheta}^{V}} \simeq \sum_i w_i \nabla_{\boldsymbol{\uptheta}^V}V^{\pi}(\mathbf{b}_{i}|\theta^{V})A_i^{\pi} $
		\EndFor
	\EndFor    

\end{algorithmic}
\label{alg:DDMAC}
\end{algorithm}
\section{Numerical experiments}\label{Sec:experim}
DDMAC inspection and maintenance policies are tested for a 9-out-of-10 system under fatigue deterioration, exploring the different statistical, structural, and cost dependencies. A second set of numerical experiments is conducted to investigate the efficiency of DDMAC policies for a 2D steel frame, also known as Zayas frame, used as a benchmark structural system for offshore engineering collapse analyses \cite{popov1980inelastic,moan1991collapse}. The numerical experiments are conducted on an Intel Core i9-7900X processor with a clock speed of 3.30 GHz. 
\subsection*{Fatigue deterioration model}
The components explored throughout the numerical investigations are assumed to be exposed to a similar fatigue deterioration, described according to the Markovian model, originally proposed in \cite{Ditlevsen2007StructuralMethods}:
\begin{equation} \label{Eq:ExamCrackGrow}
d_{t+1} =\bigg[ \Big(1-\frac{m}{2}\Big) C_{FM}S_{R}^m\pi ^{m/2}n + d_t^{1-m/2}\bigg] ^{2/(2-m)}
\end{equation}
where the crack depth, $d$, evolution over time, $t$, follows a linear-elastic fracture mechanics law with material parameters $C_{FM}$ and $m$, stress range $S_{R}$, and $n$ annual stress cycles. At the component level, failure occurs if the crack depth, $d$, exceeds a critical size, $d_c$, that corresponds to the plate thickness. In a stochastic environment, the initial crack depth, $d_0$, along with fracture mechanics model parameters are either represented by random variables or deterministic parameters as listed in Table \ref{Tab:exam1par}. 
\begin{table}[h!]
\caption{Random variables and deterministic parameters utilized to model the fatigue deterioration of the components in the numerical experiments.}\label{Tab:exam1par}
\begin{tabular}{llll}
\toprule
Variable & Distribution & Mean & SD\\
\midrule
$ln(C_{FM})$ & Normal & $-35.2$ & $0.5$ \\
$S_{R}(N/mm^{2})$ & Normal & $70$ & $10$ \\
$d_0(mm)$ & Exponential & $1$ & $1$ \\
$m$ & Deterministic & $3.5$ & - \\
$n(cycles)$ & Deterministic & $10^6$ & - \\
$t_N(yr)$ & Deterministic & $30$ & - \\
$d_c(mm)$ & Deterministic & $20$ & - \\
\bottomrule
\end{tabular}
\end{table}

\noindent The failure probability $p_{F_t}$, defined as $p_{F_t}=Pr[g_t \leq 0]$, can be computed following, for instance, a through-thickness failure criterion \cite{hlaing2020effect} by formulating the failure limit state at time step $t$ as:
\begin{equation} \label{Eq:fatLS}
g_{t}=d_c-d_t
\end{equation}

The fatigue deterioration is encoded in a deterioration rate DBN model, and ultimately shaping a factored POMDP, as shown on the left side of Fig. \ref{FIG:HDDMAC}, and presented in Section \ref{Sec:factPOMDP}. The continuous crack depth, $d$, is adequately discretizated into $|\mathcal{S}_d|=30$ states conditional on $|\mathcal{S}_{\tau}|=31$ fully observable deterioration rates states. The intervals and state space utilized for this deterioration rate model are listed in Table \ref{Tab:discrExam}. 
\begin{table}
\caption{Description of the discretization scheme implemented for the factored deterioration rate POMDP.}\label{Tab:discrExam}
\begin{tabular}{cl}
\toprule
Variable & Interval boundaries\\
\midrule
\multicolumn{2}{l}{\textbf{Deterioration rate model}} \\
$\mathcal{S}_d$ & $0, \mathrm{exp}\Big\{ \mathrm{ln}(10^{-4}):\dfrac{\mathrm{ln}(d_{c})-\mathrm{ln}(10^{-4})}{28}:\mathrm{ln}(d_{c})\Big\},\infty $ \\
$\mathcal{S}_{\tau}$ & $0:1:30$ \\
\bottomrule
\end{tabular}
\end{table}
In terms of observation model, the inspection quality is quantified with a Probability of Detection curve $PoD(d) \sim Exp[\mu = 8]$. Further details on the fatigue deterioration or observation model, including an extensive investigation of the discretization scheme can be found in \cite{morato2022optimal}.
 
\subsection{I\&M planning for a 9-out-10 system}\label{subSec:examp1}
The system explored in this application is composed of ten components, each of which is subjected to a non-stationary fatigue deterioration, as described earlier in this Section. The system is assumed to be functional if at least 9-out-of-10 components are operational (not failed), thus characterized with a single step change in terms of redundancy with respect to a series system, which would correspond to the case of a 10-out-of-10 system. The system failure probability $p_{F_{sys}}$ is efficiently computed here, as a function of the failure state of all components, by following the recursive method proposed in \cite{barlow1984computing}. 

\subsection*{Description of the I\&M decision problem} 
\begin{figure}
	\centering
		\includegraphics[scale=1]{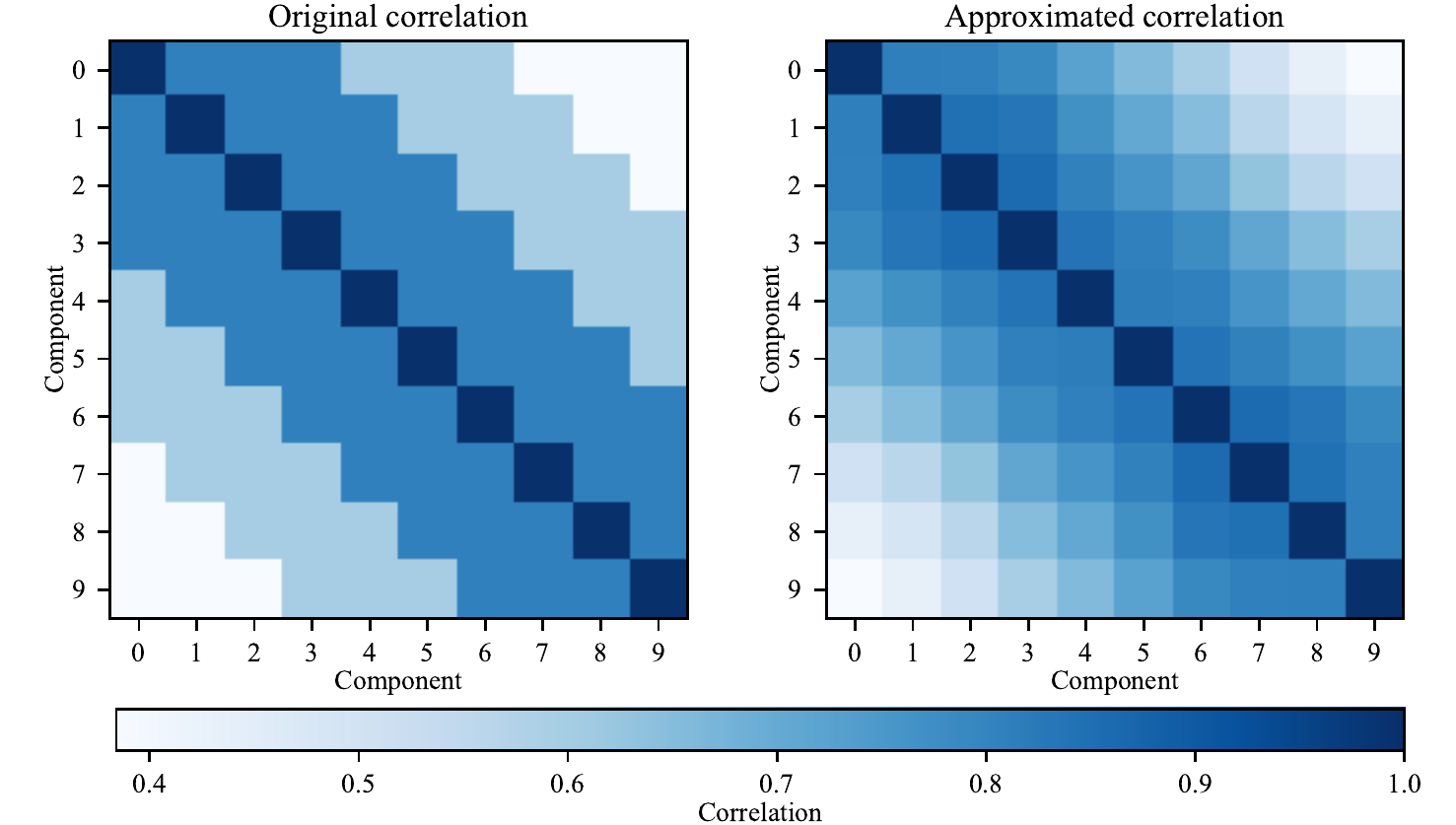}
	\caption{Representation of the initial crack size dependence among the components of the unequally correlated 9-out-of-10 system. The original deterioration correlation is represented by the colored matrix on the left. The approximated correlation structure, resulting from the derived Gaussian hierarchical model with two hyperparameters, is displayed on the right colored matrix.} 
	\label{FIG:corrDet}
\end{figure}

A total of eight I\&M planning scenarios are investigated, exploring different deterioration, risk, and cost dependencies among components. In terms of deterioration dependence, some environments are specified with an equally correlated initial crack size, $d_0$, among components, defined by an equal correlation $\rho_{eq}=0$, $\rho_{eq}=0.4$ and $\rho_{eq}=0.8$, respectively. Additionally, a deterioration environment is examined with an unequally correlated $\rho_{uq}$ initial crack size, $d_0$, among components. The unequal deterioration dependence case is originally specified with a different correlation among components of either $\rho=0.4$, $\rho=0.6$, or $\rho=0.8$, as shown on the left side of Fig. \ref{FIG:corrDet}. After a Gaussian hierarchical structure with two hyperparameters is optimized, by computing the $\boldsymbol{\lambda}$ parameters with the objective of satisfying Eq. \ref{Eq:hypCorr}, an approximated correlation structure is obtained with relatively small errors, as shown on the right side of Fig. \ref{FIG:corrDet}. The approximated correlation structure with two hyperparameters is deemed to be sufficiently accurate for the conducted experiments. Otherwise, a more accurate correlation structure can be achieved by adding more hyperparameters, at the expense of additional computational cost, as explained in Section \ref{Sec:sysEff}. For each of the aforementioned environments, specified with different deterioration dependencies, two I\&M cost models are further investigated, i.e., an I\&M cost model that incurs inspection and repair costs individually; and an I\&M cost model in which an initial campaign cost exists, if at least one component is inspected or repaired, plus a cost surplus per inspected or repaired component.

Since each component, herein denoted as fatigue hotspot, contains 930 states, defined by the joint space of 30 crack states $\mathcal{S}_d$ and 31 deterioration rate states $\mathcal{S}_{\tau}$, these are a total of 9,300 states in the experiments that do not consider deterioration correlation ($\rho_{eq}=0$). For experiments under equal correlation ($\rho_{eq}$) the total states become 744,080, rising up to $5.95\cdot 10^7$ for deterioration environments under unequal correlation ($\rho_{uq}$). The increase of the state space corresponds to the incorporation of the Gaussian hierarchical model, in which crack and deterioration rate states are formulated conditional on the hyperparameter(s) states. When the deterioration correlation is modeled equally for all components, only one hyperparameter is sufficient to satisfy Eq. \ref{Eq:hypCorr}, while two hyperparameters are added for the case of unequal correlation, as explained earlier. Each hyperparameter is discretized into 80 states, initially prescribed with equal probability for each state. Note here the importance of optimizing the number of hyperparameters included in the model, as the state space grows exponentially with the number of considered random variables. By formulating the POMDPs' transition model as dynamic Bayesian networks, the dimensionality is reduced from $|\mathcal{S}_d|^2|\mathcal{S}_{\tau}|^2$, in a flat structure, to $|\mathcal{S}_d|^2|\mathcal{S}_{\tau}|+|\mathcal{S}_{\tau}|^2$ for the uncorrelated scenario. In that case, the transition model of only one component is reduced from 864,900 to 28,861 elements. Moreover, the formulation of the environment through a hierarchical deterioration dependence model importantly enables the decoupling of the joint state space at the system level, which would grow exponentially for a flat POMDP structure, but instead grows linearly now. For instance, in the setting under unequal deterioration dependence, the joint space would be described by $\{|\mathcal{S}_d||\mathcal{S}_{\tau}|\}^{N_C}$, equaling $930^{10}$ states, while it is now instead defined by $\{|\mathcal{S}_d||\mathcal{S}_{\tau}|{N_C}|\mathcal{S}_{\alpha}||\mathcal{S}_{\beta}|+|\mathcal{S}_{\alpha}|+|\mathcal{S}_{\beta}|\}$, thus resulting in $930\cdot10\cdot80^2+80\cdot2 \simeq 5.95\cdot10^7$ states in the hierarchical model, with two hyperparameters ($\alpha$ and $\beta$) discretized into 80 states.

In terms of the neural networks' architecture, DDMAC is laid out in this application with two hidden fully-connected layers of 100 neurons for each actor network, and two hidden fully-connected layers with 200 neurons for the critic network. The learning rate is adjusted during the training of the networks from $10^{-4}$ to $10^{-5}$ for the actor, and from $10^{-3}$ to $10^{-4}$ for the critic. The exploration is set up initially with a 100\% random noise, decreasing linearly over the first 20,000 episodes to a random noise of 1\%, held constant for the remaining episodes. A more stable and efficient training was found when a prioritization of do-nothing actions is implemented at the beginning of the training, because this allows visitation of more states, thus better exploration.  
 
Following typical fatigue I\&M planning settings, inspection and repair decisions are combined into three available actions per component:  \textit{do-nothing / no-inspection}, \textit{do-nothing / inspection}, and \textit{perfect repair / no-inspection}. The action \textit{perfect-repair / inspection} is considered a priori suboptimal, without loss of generality, as it would be unusual to plan an inspection just after a component returns to its initial state. Inspections provide binary indications, i.e., \textit{detection} or \textit{no-detection} of a crack according to the observation model. In terms of costs, two different scenarios are considered. In the first case, inspection and repair costs are incurred independently per component, i.e., $r_{ins}=-1$ and $r_{rep}=-20$, respectively. In the second case, a campaign cost of $r_{camp}=-5$ is incurred if at least one component is inspected or repaired, plus a surplus per inspected or repair component of $r_{ins}=-0.2$ and $r_{per}=-20$ money units, respectively. The consequence of a system structural failure is $r_{F}=-10,000$ money units for both cases, and the discount factor $\gamma$ is 0.95 in all the experiments. 

In order to verify the optimality of the obtained DDMAC policies, predefined heuristic decision rules, adopted from \cite{LuqueDBN2019}, are optimized and compared against the results provided by DDMAC strategies. The investigated heuristic-based policies are dictated by (i) the interval between equidistant inspections $\Delta_{ins}$, (ii) how many components $n_{ins}$ are inspected at each campaign, in which the $n_{ins}$ components with higher failure probability $p_F$ are prioritized, and (iii) a perfect repair action is undertaken after a crack is detected. Initially, all the combinations of heuristics, i.e., interval between inspections $\Delta_{ins}$ and number of components inspected per campaign $n_{ins}$, are evaluated over 3,000 policy realizations. Then, the 5 sets of heuristic rules that yielded the minimum expected costs are evaluated again, this time over 10,000 policy realizations, and at the end, the set of heuristics that minimized the expected total costs are selected for comparison against DDMAC-based policies, also evaluated over 10,000 policy realizations. The resulting set of optimized heuristics is listed in Appendix A (Table \ref{Tab:app2Heur}).  
\subsection*{Results and discussion}
\begin{figure}[H]
	\centering
		\includegraphics[scale=1]{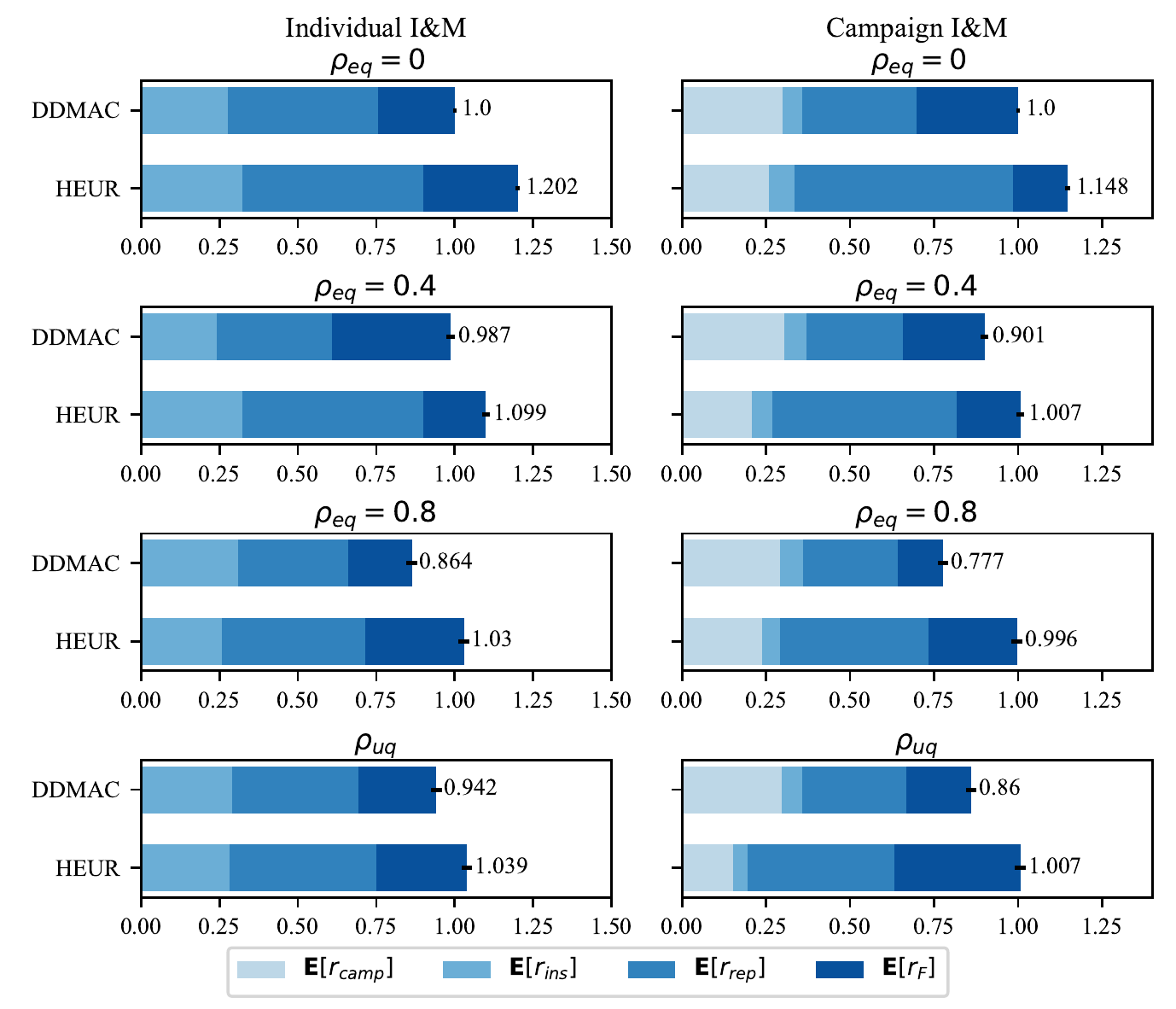}
	\caption{Expected cost results of all the numerical experiments conducted for the 9-out-of-10 system, divided into campaign $\mathbf{E}[r_{camp}]$, inspection $\mathbf{E}[r_{ins}]$, perfect-repair $\mathbf{E}[r_{rep}]$ and failure $\mathbf{E}[r_{F}]$ expected costs. On the left, DDMAC and heuristic policies, specified with an I\&M cost model, are compared for different deterioration correlation environments. Likewise, on the right, DDMAC and heuristic policies are compared for different levels of deterioration dependence, yet specified with a campaign I\&M cost model.} 
	\label{FIG:resSys1}
\end{figure}

The life-cycle expected costs obtained by evaluating the investigated policies are displayed in Fig. \ref{FIG:resSys1}, sorted in two main categories according to the specified cost model, comparing DDMAC and optimized heuristic policies and investigating the effect of adding campaign I\&M costs. For each category, four degrees of deterioration correlation are compared, i.e., no correlation ($\rho_{eq}=0$), equal correlation with ($\rho_{eq}=0.4$), equal correlation with ($\rho_{eq}=0.8$) and unequal correlation ($\rho_{uq}$). In all explored numerical examples, DDMAC outperforms the optimized heuristics, yielding life-cycle cost reductions ranging from 9.7\% to 21.9\%. The difference is more predominant for the case in which inspections and repairs are planned separately because the explored heuristic decision rules plan inspections for a group of $n_C$ components, being thus more tailored to the campaign I\&M setting. A closer examination reveals that DDMAC policy provides lower inspection, repair, and failure expected costs, with respect to heuristics-based policies, for the uncorrelated deterioration experiment specified with the individual I\&M cost model. In this case, the savings on repairs are more significant probably because the heuristic policy prescribes a repair anytime a crack detection is observed, while DDMAC-based policy usually requires more evidence than a single detection instance. 

\begin{figure}[H]
	\centering
		\includegraphics[scale=1]{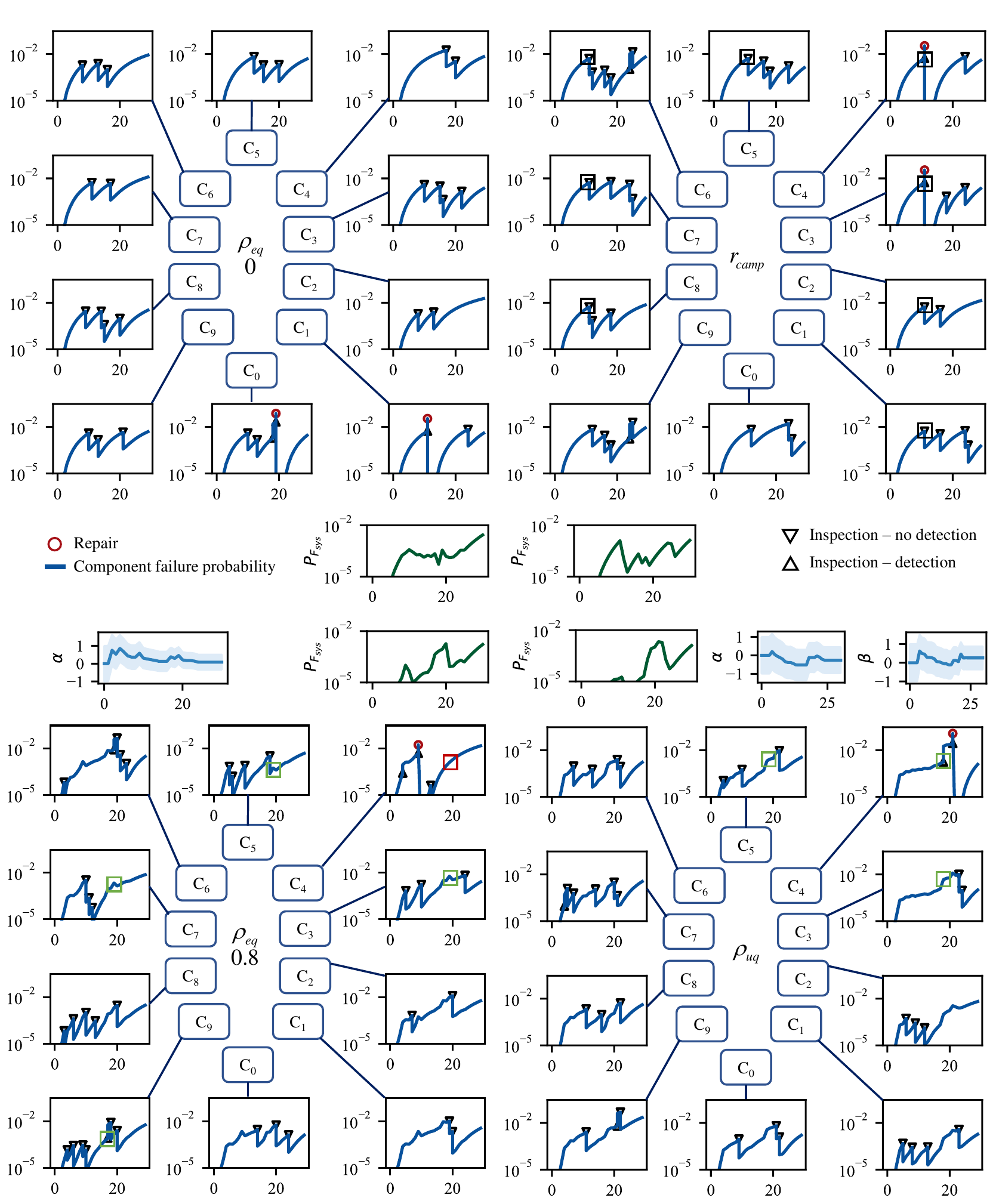}
	\caption{9-out-of-10 system policy realizations: (Upper-left) uncorrelated deterioration and individual cost model; (Upper-right) uncorrelated deterioration and campaign cost model; (Lower-left) equally correlated environment ($\rho_{eq}=0.8$) and individual cost model; (Lower-right) unequally correlated deterioration and individual cost model. Failure probabilities at the component level are depicted by blue lines, inspection indications are represented by upwards (detection) or downwards (no-detection) triangles and repairs are circled in red. At the system level, the failure probability is represented by green diagrams and the evolution of the hyperparameters, under correlated deterioration, is described by light-blue graphs. As mentioned in the text, specific policy and inference effects are marked with coloured rectangles.}
	\label{FIG:polSys1}
\end{figure}

With regard to deterioration dependence, highly correlated environments result generally in lower expected total cost, as observed in Fig. \ref{FIG:resSys1}. Information collected on one component, in environments under deterioration correlation, also provides information to other components' states in the system. One can clearly observe the effect of inspection and repairs on the resulting component and system failure probabilities as well as the oscillation of the hyperparameters around an uncorrelated mean of 0 when inspections among components are not consistent, i.e., some inspections indicate damage and some others do not, whereas the opposite is observed when all inspections are consistent, e.g., no damage indicators. For instance, a crack detection observed on component 9 for the case of equal correlation $\rho_{eq}=0.8$, leads to an incremental increase on the failure probability of other non-repaired components, as indicated inside the green rectangles on the lower-left corner of Fig. \ref{FIG:polSys1}. This effect can be also visualized when observing the impact of a crack detection on component 4, for the case under unequal deterioration dependence, marked by green rectangles on the lower-right plot of Fig. \ref{FIG:polSys1}. In this case, components 3 and 5, highly correlated with component 4, as indicated in Fig. \ref{FIG:corrDet}, are clearly affected by the observed crack detection. 
\begin{figure}[H]
	\centering
		\includegraphics[scale=1]{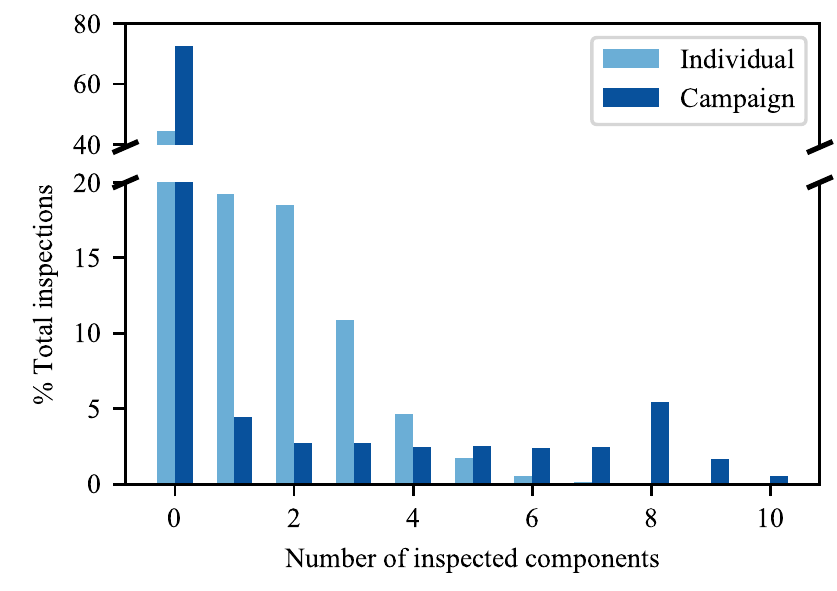}
	\caption{Comparison of DDMAC policies specified with either individual (light blue) or campaign (dark blue) I\&M cost models. For each case, the number of inspected components per time step is represented in a histogram based on 10,000 policy realizations.}
	\label{FIG:insDepend}
\end{figure}

Moreover, highly correlated deteriorating environments induce higher variability on the expected total costs, as shown by the black error bars in Fig. \ref{FIG:resSys1}. The variability can be attributed to the very different resulting policy paths following inspections, depending on whether the collected observations indicate cracks or not. If a crack is detected on one component, the other components' failure probabilities increase, and repair actions or additional inspections will be planned thereafter. Conversely, if a crack is not detected, the failure probability of all the correlated components will decrease, inducing less repair actions in the future. Interestingly, policies under dependent environments do not always plan fewer inspections, as it could be expected due to the additional information gained through the underlying correlation among components, but instead highly correlated environments might plan more inspections, often resulting in significant failure risk reductions, as displayed for the case with $\rho_{eq}=0.8$ in Fig. \ref{FIG:resSys1}. The effect of repair actions on the deterioration dependence structure can also be clearly observed. Once an element is repaired, its damage belief becomes independent from the global hyperparameter(s), and thus inspection outcomes from other components do not influence the repaired component, and vice versa. As illustrated with a red rectangle at the lower-left corner of Fig. \ref{FIG:polSys1} (i.e., equal correlation $\rho_{eq}=0.8$ setting), after component 4 is repaired, its failure probability is not influenced now by inspection results retrieved from other components, e.g., a crack detection observed on component 9.

To further investigate the effect of including campaign utilities within the cost model, a histogram over 10,000 policy realizations is shown in Fig. \ref{FIG:insDepend} for a DDMAC policy in which inspections and repairs are incurred separately (light blue) and another DDMAC policy considering the expense of campaign actions (dark blue). The deterioration environment for both DDMAC policies does not consider deterioration correlation among components. The emphasis of Fig. \ref{FIG:insDepend} is on the number of components inspected on every occasion an inspection is planned. If inspections and repairs are paid individually for each component, information on only one or two components usually suffices, whereas inspection of more than five components per time step is rare. In contrast, eight-component inspections become the predominant inspection decision if an initial campaign cost is included in the cost model. This system effect can be visualized, for example, in the policy realizations shown at the top of Fig. \ref{FIG:polSys1}, in which black rectangle-marked components are inspected at the same year for the DDMAC policy under campaign costs. The policy for the campaign cost model, therefore, tends to group inspection and repair actions at the same year, avoiding, if possible, unnecessary campaign costs associated with one or two inspected components. In some cases, campaigns are however planned for only one or two components, contrasting with the static inspection decision rules imposed by heuristics, where a specific number of inspections is fixed for all the inspection campaigns. Based on the above, we observe that DDMAC is able to devise dynamic I\&M policies according to the specified cost model, whether under campaign costs or individual inspection and repairs, and to provide an advanced, flexible, and adaptive decision-making framework.       

\subsection{I\&M planning for Zayas frame}
In the first set of numerical experiments, conducted for a 9-out-of-10 system, the focus was mainly directed to the investigation of the deterioration dependence among identical components and the effects of including a campaign cost within the cost model. In this second application, we further explore how I\&M DDMAC policies are able to inherently capture the relative importance of each element with respect to the system structural reliability. The structural system of study, in this case, is the 2-dimensional Zayas frame, well studied in many structural reliability analysis applications \cite{popov1980inelastic,moan1991collapse,Schneider2017a}. Zayas frame is composed of two columns, which along with 13 braces, sustain a rigid beam at the top. The geometry and material properties used in this work are presented in the Appendix B (Fig. \ref{FIG:Zayasgeom}) and are the same as the ones used in \cite{Schneider2017a}.  
\subsection*{Description of the I\&M decision problem}
In this application, DDMAC policies are identified for two I\&M settings: (i) under equal deterioration dependence among components with $\rho_{eq}=0.4$, and (ii) assuming independence among components' deterioration. The state space for the latter includes 30 crack states along with 31 deterioration rate states, for each of the 22 hotspots (i.e., components), resulting therefore in a total of 20,460 input variables; while the input variables for the former climbs to approximately $1.6\cdot 10^6$ states, including 80 states for the one discretized hyperparameter. The benefits associated with the proposed decoupled hierarchical structure are very significant, since the state space has a dimension of $930^{22}$ if the joint states of all hotspots are explicitly considered.

Similarly to the experiments reported in Section \ref{subSec:examp1}, the decision maker is here able to select three actions per hotspot at each time step: \textit{do-nothing / no-inspection}, \textit{do-nothing / inspection}, and \textit{perfect repair / no-inspection}. Again, inspections provide binary crack indications (detected or no-detected), equally modeled for each component by the observation model described in the beginning of this Section. As for the cost model, inspections and repairs (planned individually for each component) cost $r_{ins}=-1$ and $r_{pr}=-15$ money units, respectively, while the system failure cost is defined as $r_{F}=-50,000$ money units. All costs are discounted to the present value by a $\gamma=0.95$ factor. DDMAC's architecture is similar to the first application, featuring two hidden fully-connected layers of 150 neurons for each actor, and two hidden layers of 300 neurons for the critic network. Learning rate, prioritization of actions and additional exploration settings are equally defined as for the first application. The investigated heuristic-based policies rely on the same set of decision rules introduced in the former numerical experiments, accounting in this case, for inspections intervals $\Delta_{ins}$ and inspected hotspots per campaign $n_C$. Both DDMAC and heuristic policies are evaluated over 30,000 episodes and the results, in terms of expected total costs, are showcased in Fig. \ref{FIG:costZayas}.
\subsection*{System failure probability}
Offshore structures are exposed to fatigue and corrosion deterioration due to the combined cyclic effect of waves and wind in a harsh marine environment. Initial defects at geometric discontinuities or at welded regions (hotspots) grow over time, becoming critical if maintenance actions are not timely undertaken. In this study, and following the experiments conducted in \cite{DBNLuque2016,Schneider2017a}, a total of 22 hotspots are considered, located at the joints at the braces or columns. Each brace is associated with either one or two hotspots, as illustrated in Fig \ref{FIG:polZay}, at critical locations for fatigue deterioration. The fatigue deterioration is assumed similar for all hotspots, modeled by the same deterioration process as for the 9-out-of-10 structural system (Section \ref{subSec:examp1}). 

The failure of the system is defined here as the incapacity of the frame to withstand the concentrated horizontal load applied at the upper-left corner. At the component level, the health of each hotspot is described by the vector $\mathbf{F_h}$, in which $F_h$ is a binary variable with $F_h=0$ indicating a hotspot failure and $F_h=1$ corresponding to a hotspot survival. The failure probability of a hotspot $p_F^{(h)}$ corresponds thus to the probability of being in state $F_{h}=0$. At an element level, the state of each brace is represented by a vector $\mathbf{x_{el}}$, considering $x_{el}=0$ if the element has failed and $x_{el}=1$ otherwise. Assuming that a brace fails if any of its associated hotspots fail, the failure probability of an element $p_{F}^{(el)}$, i.e., $Pr(x_{el}=0)$, can be therefore computed as a series system:
\begin{equation}\label{Eq:elemHots}
p_{F}^{(el)}= 1- \prod_{h\in N_{h}} \left[1-p_F^{(h)}\right]
\end{equation}

At the system scale, the health of the frame depends on the state of all its constitutive elements, i.e., 13 braces, and the failure probability of the system $p_{F_{sys}}$ is computed herein as a function of all the element state combinations. A total of 8,192 ($=2^{13}$) non-linear static push-over simulations have been run with the assistance of the computer code `USFOS' (available within the software package Sesam) \cite{soreide1993usfos}, before the training of DDMAC, so that the failure probability of the system conditional on all element state combinations is explicitly and directly defined. The element configuration for each push-over simulation is arranged according to the element state vector $\mathbf{x_{el}}$, removing the braces associated with a failed state $x_{el}=0$. The resistance $L_{col}(\mathbf{x_{el}})$ of each element state combination cases is retrieved from the conducted push-over simulations.     

The collapse event of the frame is defined as the probability of the external horizontal load exceeding the structural system resistance $Pr(L>L_{col})$. In this case, the horizontal load is modeled as a lognormal random variable with mean $\mu_L=70$kN and $25\%$ coefficient of variation, while no uncertainty is associated with the resistance, a reasonable assumption when the external load is highly uncertain in comparison with the resistance \cite{Schneider2017a}. The failure probability of the system, $p_{F{sys}}$, conditional on the element state vector $\mathbf{x_{el}}$ can be then defined directly from the probability density function of the load, $f_L$,:
\begin{equation}\label{Eq:sysPFCond}
p_{F{sys}}^{(\mathbf{x_{el})}} = \int_{L_{col}}^\infty f_L(x) dx
\end{equation}
In the undamaged case, i.e., no elements are removed from the original configuration, the collapse load is 247 kN, resulting in a failure probability of approximately $10^{-4}$. The state of the frame is, however, computed conditional on the state of all the elements, and to do so the probability of being in each state combination should be computed. We follow the iterative procedure proposed in \cite{song2009system} to compute the probability of being in each element state $\mathbf{q}\doteq p(\mathbf{x_{el}})$ as a function of the element failure probability $p_F^{(el)}$ and the element survival probability $\bar{p}_F^{(el)}$:
\begin{equation}\label{Eq:stateSyst}
\begin{split}
\mathbf{q_{[1]}} = \begin{bmatrix}p_F^{(1)} & \bar{p}_F^{(1)}\end{bmatrix}^T \\
\mathbf{q_{[i]}} = \begin{bmatrix} \mathbf{q_{[i-1]}}\cdot p_F^{(i)} \\
\mathbf{q_{[i-1]}}\cdot \bar{p}_F^{(i)}\end{bmatrix}
\end{split}
\end{equation}
Finally, the system failure probability $p_{F{sys}}$ is equal to the system failure probability conditional on the element state $p_{F{sys}}^{(\mathbf{\mathbf{x_{el}}})}$ multiplied by the probability of being in that state $q^{(\mathbf{x_{el}})}$:
\begin{equation}\label{Eq:sysPF}
p_{F{sys}} = \sum_{\mathbf{\mathbf{x_{el}}} \in \mathcal{X}_{el}} \Big[ p_{F{sys}}^{(\mathbf{x_{el}})}\cdot q^{(\mathbf{x_{el}})} \Big]
\end{equation}
\pagebreak
\subsection*{Results and discussion}
\begin{figure}
	\centering
		\includegraphics[scale=1]{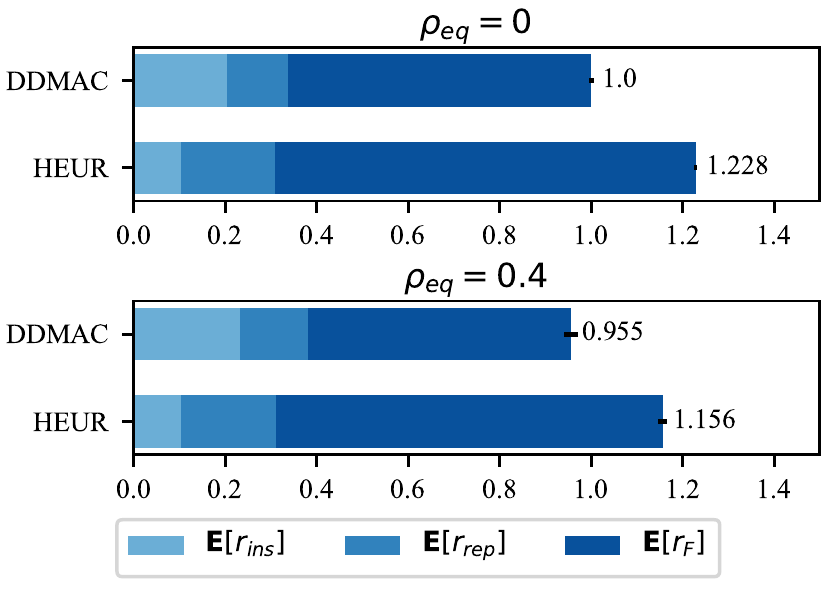}
	\caption{Expected  cost  results  for  the  numerical  experiments  conducted  for  the  Zayas frame,  divided into inspection $\mathbf{E}[r_{ins}]$, perfect-repair $\mathbf{E}[r_{rep}]$ and failure $\mathbf{E}[r_{F}]$ expected costs. DDMAC and heuristics are compared under an uncorrelated deterioration environment at the top, and under an equally correlated environment ($\rho_{eq}=0.4$) at the bottom.}
	\label{FIG:costZayas}
\end{figure}

\begin{figure}[H]
	\centering
		\includegraphics[scale=1]{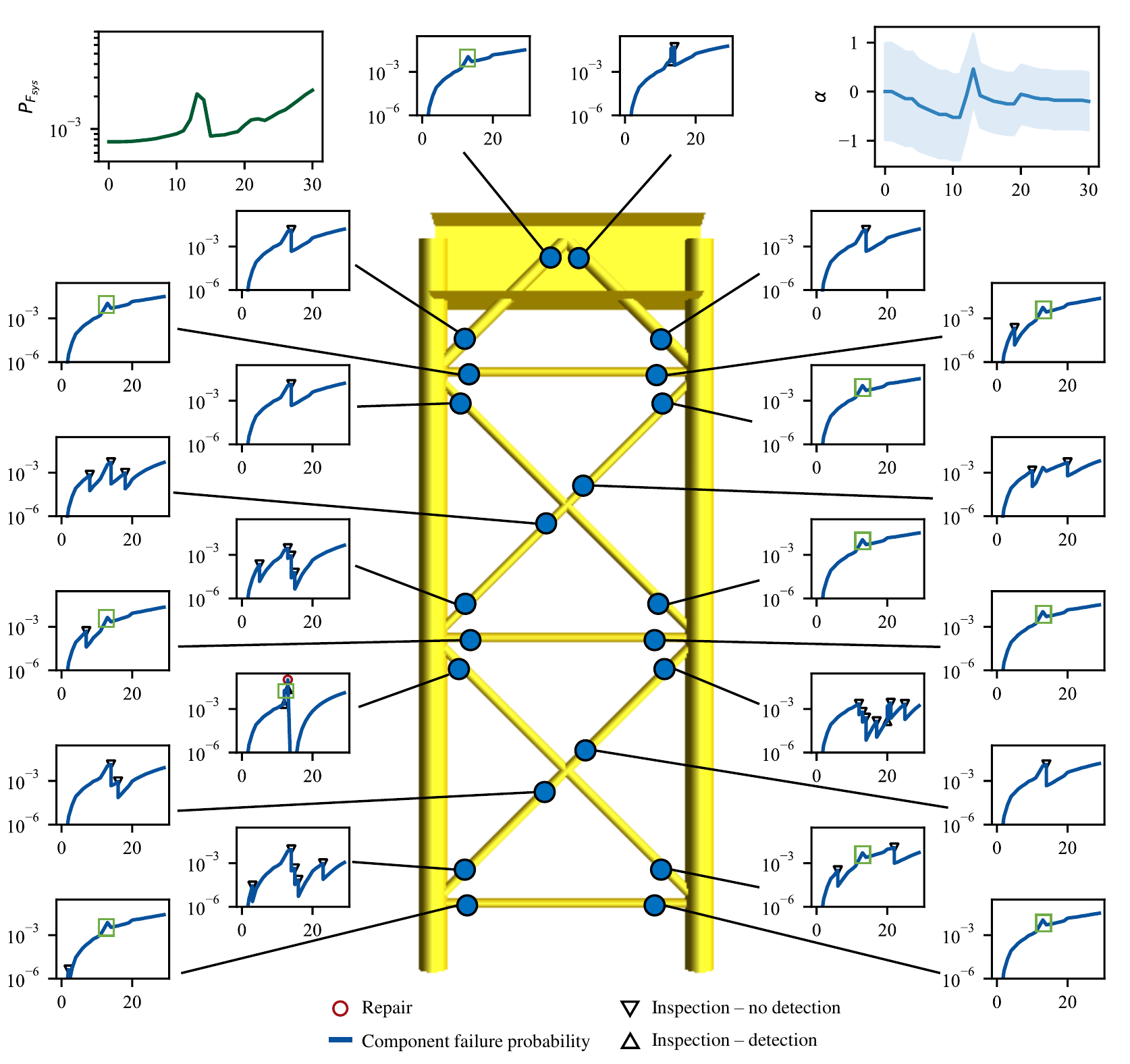}
	\caption{Zayas frame sample policy realization in an equal deterioration correlation environment ($\rho_{eq}=0.4$). The failure probability of each hotspot is depicted by a blue line, inspection indications are represented by upwards (detection) or downwards (no-detection) triangles, and perfect repairs are circled in red. At the system level, the failure probability, at the top-left corner, and system-effects, as these are explained in the main text, are represented by a green line and squares, respectively. The evolution of the hyperparameters over time is plotted in a light blue diagram, at the top-right corner.}
	\label{FIG:polZay}
\end{figure}

The comparison between DDMAC and optimized heuristics follows the same trend as that of the 9-out-10 structural system experiments. In terms of expected life-cycle costs, DDMAC policies outperform heuristic-based policies in the two tested settings, as shown in Fig. \ref{FIG:costZayas}, with cost savings ranging from 20.1\% to 22.8\%. A slight decrease in the expected life-cycle costs can also be observed for the case under deterioration correlation, as a result of the reduction of failure risk. This is expected, since under deterioration dependence, i.e., the initial crack size among the hotspots is correlated, an observation collected at one hotspot also provides information to other hotspots, accordingly updating their damage state belief. These beliefs are updated for both detection and no detection observation outcomes, as illustrated in Fig. \ref{FIG:polZay}. At year 12, a crack is detected at the lower X-brace, and this observation shows up as a failure probability update for the other components, marked with a green rectangle in the plots, an effect that can also be observed clearly in the updated mean of the hyperparameter, $\alpha$. In most policy realizations and hotspot inspections, however, the most likely observation outcome is no-detection, explaining the observed risk reduction in cases under deterioration correlation. The effect of deterioration dependencies among components on the resulting system failure probability can also be visualized in Fig. \ref{FIG:polZay}, e.g., a crack detection at hotspot 6 ultimately induces a kink in the system failure probability around year 20.  

\begin{figure}[H]
	\centering
		\includegraphics[scale=1]{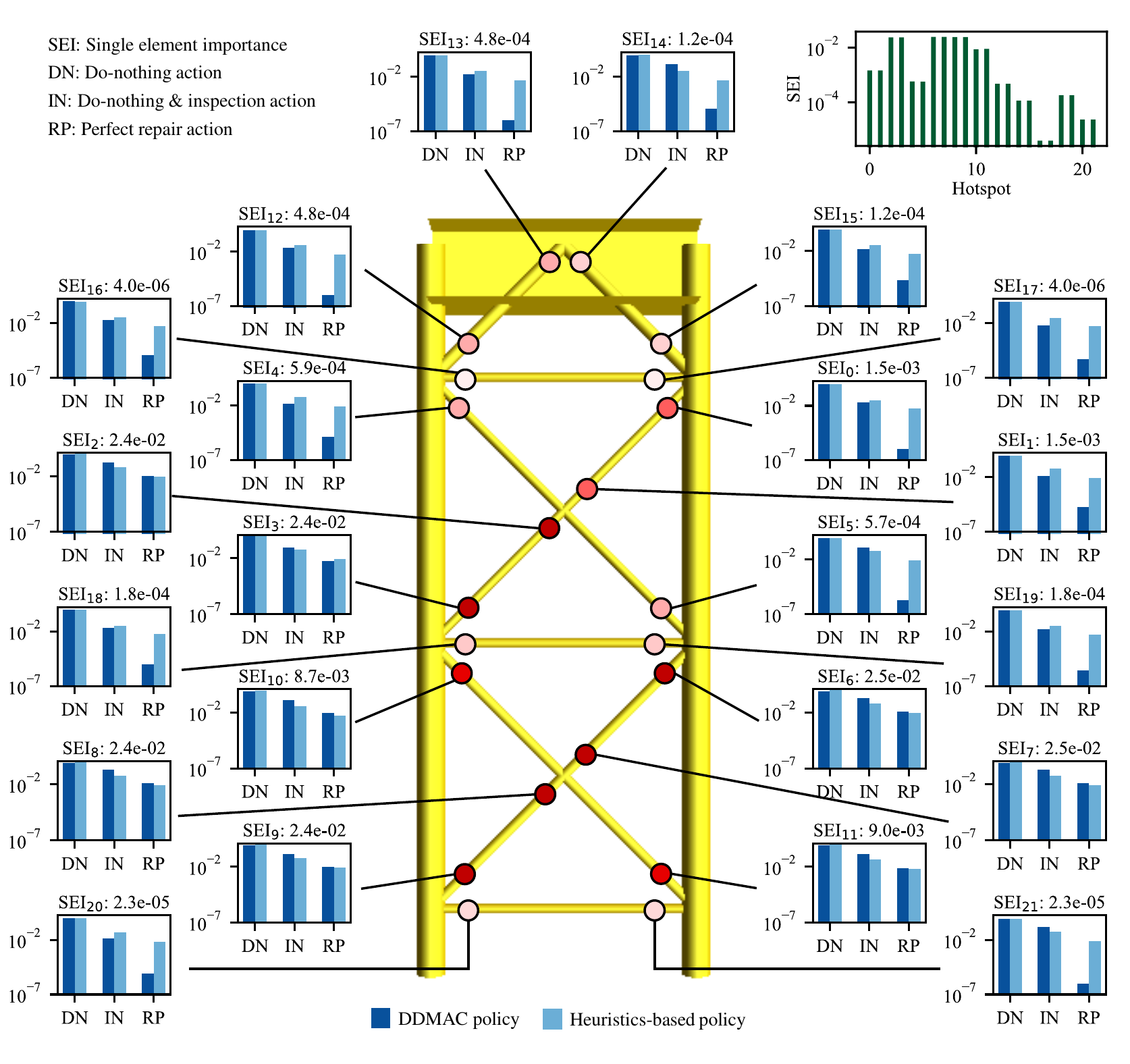}
	\caption{Histograms of DDMAC and heuristics-based policy actions of 30,000 realizations of this Zayas frame setting for a 30-year policy period. The Single Element Importance metric (SEI) associated with each fatigue hotspot is indicated at the top of each histogram and summarized at the green top-right diagram. The relative importance of each hotspot is also represented by color, with a darker red being a more critical element for the structural reliability of the system.}
	\label{FIG:SEIZay}
\end{figure}

Essentially, DDMAC is able to discover the importance of each hotspot regarding the structural reliability of the frame. To explore this system effect, the Single Element Importance (SEI) measure is calculated for each hotspot. The concept of SEI, as defined in \cite{straub2011reliability}, determines the importance of each element to the system structural reliability by subtracting the undamaged system failure probability $p_{F_{sys}}$ from the system failure probability with the element removed ($\sim el$). In this case, and since each element is defined as a series system of hotspots, the SEI can be directly computed for each hotspot $h$, determining each hotspot importance as: 
\begin{equation}\label{Eq:SEIdef}
SEI_{h} = p_{F_{sys}}^{(\sim h)} - p_{F_{sys}}
\end{equation}
The SEI of a vital element for the structural system is thus higher than the SEI of a less important component. The structural element importance (SEI) of each hotspot is shown in Fig. \ref{FIG:SEIZay}, along with histograms of the actions taken at each component during 30,000 DDMAC (dark blue) and optimized-heuristic (light blue) realizations. As represented by the dark green bar diagram at the top-right corner of Fig. \ref{FIG:SEIZay}, and in agreement with the findings reported in \cite{Schneider2017a}, the critical hotspots are located at the X-braces, whereas the less critical hotspots are the ones connecting the horizontal braces. While the \textit{do-nothing} action is dominant and \textit{inspection} actions are distributed similarly among components, the distribution of \textit{repair} actions among hotspots differs for DDMAC and heuristics-based decision rules. DDMAC plans repairs mainly for important elements with respect to the global structural reliability, i.e., with a high SEI, such as hotspots 6 and 7, whereas less important components for the system structural reliability are less frequently repaired. In contrast, the heuristic-based policy plan component repairs nearly evenly, disregarding the influence of each hotspot to the reliability of the system. We can therefore conclude that DDMAC policies are able to inherently identify the system effects attributed to the structural and reliability importance of each element for the entire system.
 
\pagebreak
\section{Concluding remarks}  \label{Sec:conclusions}
This paper introduces an efficient algorithmic framework for inference and optimal decision-making under uncertainty for engineering systems exposed to deteriorating environments. In terms of inference, a Gaussian hierarchical structure is presented, within a dynamic Bayesian network model, and further formalized here with the objective of enabling the treatment of engineering systems under general, unequal deterioration correlation settings, considering also the effect of maintenance actions. The proposed efficient inference framework is then seamlessly integrated with principled optimization methods by formulating the decision-making problem as a factored Partially Observable Markov Decision Process (POMDP), with its dynamics encoded by Bayesian network conditional structures. The system life-cycle realizations collected by simulating the specified POMDP are employed by a multi-agent actor-critic deep reinforcement learning algorithm, able to identify optimal strategies in very high dimensional state, action, and observation spaces, commonly found in practical structural and engineering systems. In particular, we demonstrate through numerical experiments that the proposed approach provides efficient inspection and maintenance (I\&M) policies, outperforming state-of-the-art policies, and enables a systematic treatment of system effects, that is autonomously and intrinsically reflected in the identified strategies.

POMDP-based policies, parametrized in high-dimensional settings through the Deep Decentralized Multi-agent Actor Critic (DDMAC) algorithm, map the current belief state of the system to a probability distribution of possible actions. These stochastic policies are thus prescribed as a function of the belief state, which is a sufficient statistic of the history of actions and observations. Constructing the policies based on a sufficient statistic feature enables more effective optimal decision-making strategies than static optimization approaches, which are constrained by the limited space explored during the policy search.  POMDP-based policies provide an additional flexibility to the decision maker, who can opt for an alternative decision at some point, for any reason, and the policy through the updated belief state will be automatically adapted thereafter, yielding near-optimal results. 

DDMAC policies are approximated by actor neural networks, whose weights are learned according to noisy rewards collected at the system level. By including deterioration dependence among components in the simulated environment, and by formulating the cost model at the system level, DDMAC policies are able to intrinsically consider the following system effects:   
\begin{itemize}
    \item In deterioration dependent environments, observing the state of one component provides indirect information to the other components of the system, modulated by their degree of correlation. In the tested I\&M planning scenarios, environments with higher correlation resulted in a reduction of expected costs, usually characterized by lower expected failure risk. As structural systems are designed according to high reliability standards, demanding a low failure risk, observations mostly indicate sound structural states, which in highly correlated environments results in a global reduction of failure risk. As opposed to independent deterioration settings, higher variability in the expected costs is observed in dependent environments, in which very different I\&M policy scenarios can be experienced based on the acquired observations. 
    \item A clustering effect on inspections and repairs is observed in settings that include a campaign cost model, i.e., a fixed cost is activated if at least one component is repaired or inspected. In this case, policies seek to avoid planning single or few inspections and repairs at one time step. Instead, inspection and maintenance actions are generally grouped, saving the additional campaign cost associated with inspecting and repairing only few components within one campaign.
    \item Maintenance actions are influenced by the relative importance of the components to the system structural reliability. As observed in the steel frame application, repairs were mostly allocated to critical elements, whereas components less important to the global reliability were less often repaired. 
\end{itemize}
In this work, the deterioration environment is formulated as a discrete state POMDP, in which exact Bayesian inference can be conducted. Further research can be focused on the proper development of continuous state POMDPs and/or on modeling, inference, and optimization procedures that would allow for further reduction of the state/action space dimensionality.
\pagebreak
\section*{Acknowledgements}
This research is funded by the National Fund for Scientific Research in Belgium F.R.I.A. - F.N.R.S. This support is gratefully acknowledged. Dr. Papakonstantinou would further like to acknowledge that this material is also based upon work supported by the U.S. National Science Foundation under Grant No. 1751941. Dr. Andriotis would like to acknowledge the support of the TU Delft AI Labs program. Finally, the authors would like to acknowledge the support provided by DNV GL Digital Solutions for granting access to the software package `Sesam'. 

\section*{Appendix A. Optimized heuristic decision rules.}\label{Sec:appendB}
\begin{table}[H]
\renewcommand\thetable{A1}
\caption{List of optimized heuristic decision rules employed in the numerical experiments. For each considered setting, the resulting heuristic decision rules dictate inspections for $n_C$ components at equidistant intervals of $\Delta_{ins}$ years. $\rho$ indicates equal ($eq$) or unequal ($uq$) deterioration correlation among components.}\label{Tab:app2Heur}
\begin{tabular}{lllll}
\toprule
Setting & Deterioration correlation & Cost model & $\Delta_{ins}$ & $n_C$\\
\midrule
9-out-of-10 system & $\rho_{eq}=0$ & Individual & $6$ & $10$ \\
9-out-of-10 system & $\rho_{eq}=0.4$ & Individual & $6$ & $10$ \\
9-out-of-10 system & $\rho_{eq}=0.8$ & Individual & $6$ & $8$ \\
9-out-of-10 system & $\rho_{uq}$ & Individual & $5$ & $7$ \\
9-out-of-10 system & $\rho_{eq}=0$ & Campaign & $5$ & $10$ \\
9-out-of-10 system & $\rho_{eq}=0.4$ & Campaign & $6$ & $10$ \\
9-out-of-10 system & $\rho_{eq}=0.8$ & Campaign & $5$ & $7$ \\
9-out-of-10 system & $\rho_{uq}$ & Campaign & $8$ & $10$ \\
Zayas frame & $\rho_{eq}=0$ & Individual & $10$ & $16$ \\
Zayas frame & $\rho_{eq}=0.4$ & Individual & $10$ & $16$ \\
\bottomrule
\end{tabular}
\end{table}

\pagebreak

\section*{Appendix B. Zayas frame geometry and material properties.}\label{Sec:appendA}
\begin{figure}[H]
\renewcommand\thefigure{B1}
	\centering
		\includegraphics[scale=1]{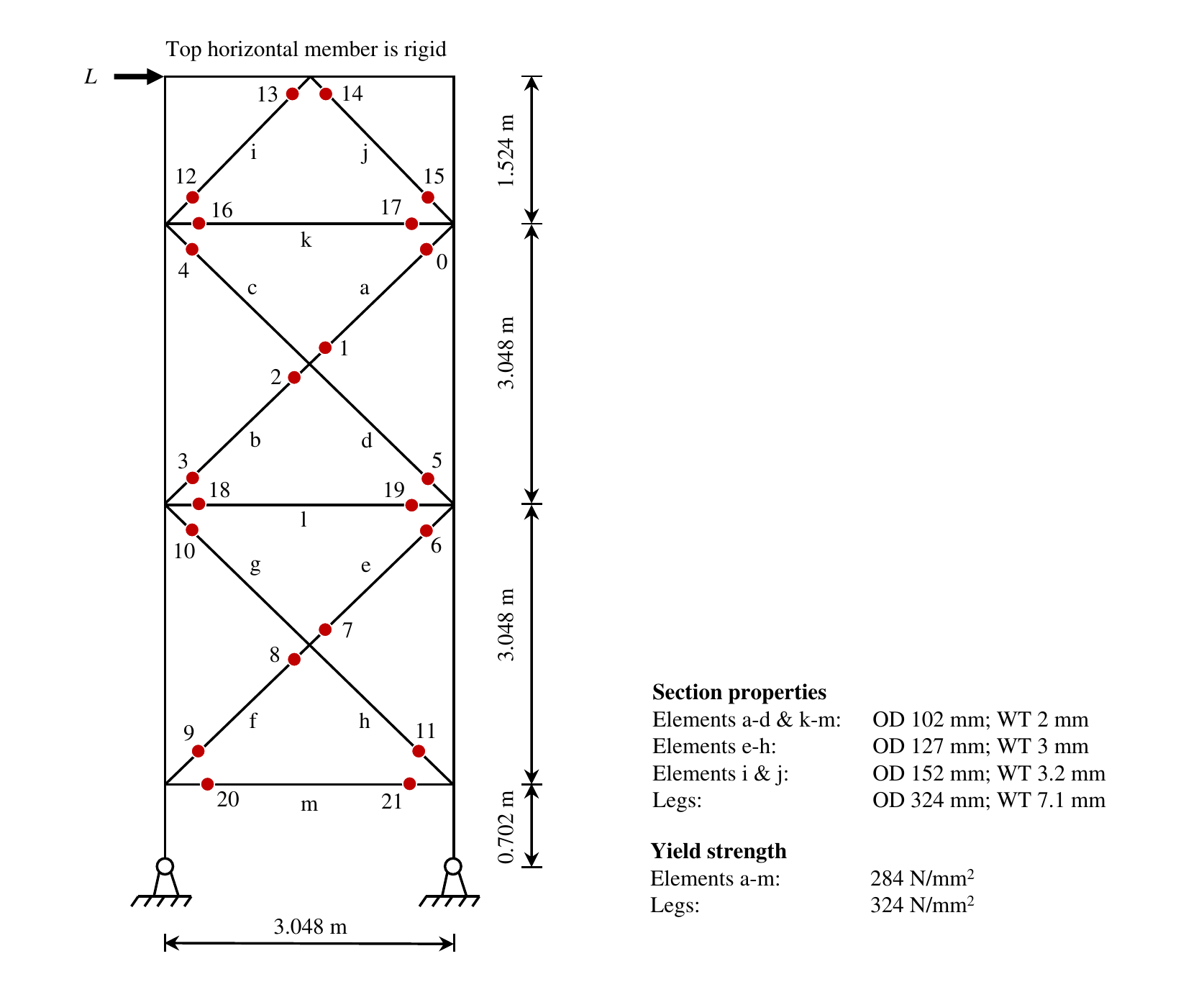}
	\caption{Zayas frame representation \cite{popov1980inelastic,Schneider2017a}. Elements are denoted with lower-case letters and fatigue hotspots are designated with numbers. The outer diameters, OD, wall thicknesses, WT, and mechanical properties corresponding to each element of the frame are specified in the diagram. An external horizontal load, $L$, is applied at the upper-left corner of the frame.}
	\label{FIG:Zayasgeom}
\end{figure}

\pagebreak



\bibliographystyle{elsarticle-num} 
\bibliography{Inference_dyn_decision_making_DRL}





\end{document}